\documentclass[10pt,acmtog]{acmart}

\usepackage{multicol,graphicx,amssymb,mathrsfs,amsmath,pifont,amscd,latexsym,color,fancyhdr,CJK,url,longtable, pifont,subfig,epstopdf,setspace,multirow,colortbl,tabularx,threeparttable, booktabs, verbatim, bbm,listings, balance,color,indentfirst,xspace,hyperref,anyfontsize,mathtools}

\DeclarePairedDelimiter{\ceil}{\lceil}{\rceil}
\DeclarePairedDelimiter\floor{\lfloor}{\rfloor}

\usepackage[linesnumbered,boxed,algosection]{algorithm2e}
\SetKwComment{tcp}{$\triangleright$ }{}

\SetCommentSty{mycommfont}

\usepackage{array}
\newcolumntype{L}[1]{>{\raggedright\let\newline\\\arraybackslash\hspace{0pt}}m{#1}}
\newcolumntype{C}[1]{>{\centering\let\newline\\\arraybackslash\hspace{0pt}}m{#1}}
\newcolumntype{R}[1]{>{\raggedleft\let\newline\\\arraybackslash\hspace{0pt}}m{#1}}

\newcommand{\framework}{DeepCache\xspace}
\newcommand{\sys}{\framework{}}
\newcommand{\shrink}{\emph{cache erosion}\xspace}
\newcommand{\variation}{scene variations\xspace}

\newcommand{\revise}[1]{{#1}}

\setcopyright{none}
\acmDOI{10.1145/3241539.3241563}
\acmISBN{none}

\usepackage{color,soul}

\definecolor{refkey}{rgb}{0,0,1}
\definecolor{labelkey}{rgb}{0,1,0}

\newcommand{\engine}{deep learning engine}

\renewcommand{\paragraph}[1]{\vskip 3pt\noindent\textbf{#1 }}	 

\begin{document}
\title{\framework: Principled Cache for Mobile Deep Vision}


 \author{Mengwei Xu}
 \affiliation{%
   \institution{Peking University, MoE, Beijing, China}
   }
   \email{xumengwei@pku.edu.cn}

  \author{Mengze Zhu}
 \affiliation{%
   \institution{Peking University, MoE, Beijing, China}
   }
   \email{zhumz@pku.edu.cn}


 \author{Yunxin Liu}
 \affiliation{%
   \institution{Microsoft Research}
   \city{Beijing, China}}
\email{yunxin.liu@microsoft.com}

 \author{Felix Xiaozhu Lin}
 \affiliation{%
   \institution{Purdue ECE}
   \city{West Lafayette, Indiana, USA}}
   \email{xzl@purdue.edu}
   
    \author{Xuanzhe Liu}\thanks{Xuanzhe Liu is the paper's corresponding author.}
 \affiliation{%
   \institution{Peking University, MoE, Beijing, China}}
\email{xzl@pku.edu.cn}


%
%


\begin{abstract}

We present DeepCache, a principled cache design for deep learning inference in continuous mobile vision.
DeepCache benefits model execution efficiency by exploiting temporal locality in input video streams. 
It addresses a key challenge raised by mobile vision: 
the cache must operate under video scene variation, 
while trading off among cacheability, overhead, and loss in model accuracy. 
At the input of a model, DeepCache discovers video temporal locality by exploiting the video's internal structure, for which it borrows proven heuristics from video compression; 
into the model, DeepCache propagates regions of reusable results by exploiting the model's internal structure. 
Notably, DeepCache eschews applying video heuristics to model internals which are not pixels but high-dimensional, difficult-to-interpret data. 


\revise{
Our implementation of DeepCache works with unmodified deep learning models, requires zero developer's manual effort, and is therefore immediately deployable on off-the-shelf  mobile devices.
Our experiments show that DeepCache saves inference execution time by 18\% on average and up to 47\%. 
DeepCache reduces system energy consumption by 20\% on average.}

\end{abstract}


\maketitle

\section{Introduction} \label{sec:intro}
With ubiquitous cameras on mobile and wearable devices, \emph{continuous mobile vision} emerges to 
enable a variety of compelling applications, including cognitive assistance~\cite{Chen:2017:ESL:3132211.3134458}, life style monitoring~\cite{computers6010004}, and street navigation~\cite{conf/sensys/ChenRDBB15}.
%
To support continuous mobile vision, Convolutional Neural Network (CNN) is recognized as the state-of-the-art algorithm: 
a software runtime, called deep learning engine, ingests a continuous stream of video images\footnote{We refer to them as a \textit{mobile video stream} in the remainder of the paper.};
for each input frame the engine executes a CNN model as a cascade of \textit{layers}, produces intermediate results called \textit{feature maps}, and outputs inference results. 
Such CNN executions are known for their high time and space complexity, stressing resource-constrained mobile devices. 
Although CNN execution can be offloaded to the cloud~\cite{conf/mobisys/HanSPAWK16,siri}, it becomes increasingly compelling to execute CNNs on device~\cite{conf/mobisys/MathurLBBFK17,conf/ipsn/LaneBGFJQK16,conf/sensys/ChenRDBB15}, which ensures fast inference, preserves user privacy, and remains unaffected by poor Internet connectivity.


\begin{figure}[t!]
\centering
\includegraphics[width=0.45\textwidth{}]{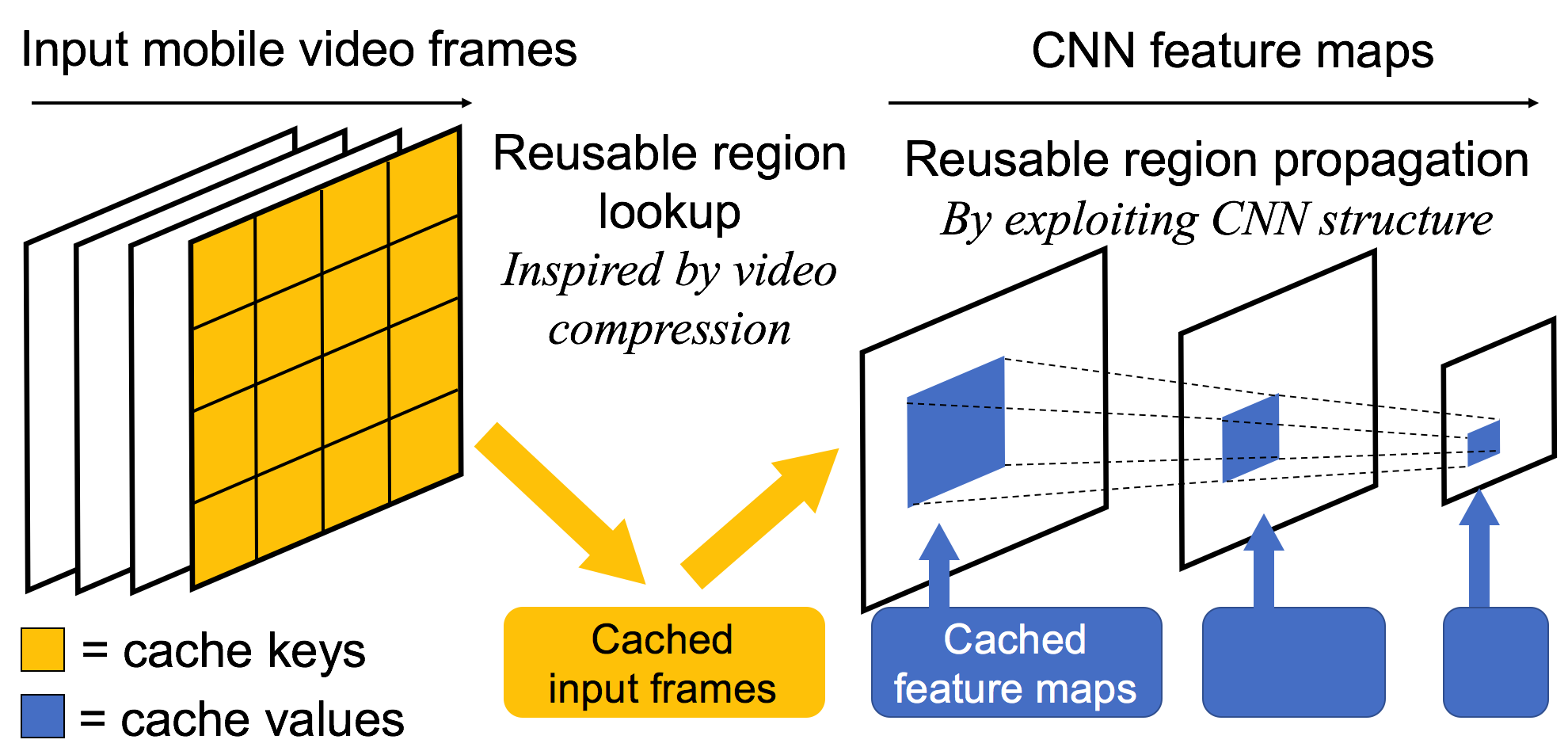} 
\caption{The overview of \framework{}.}
\label{fig:overview}
\end{figure}

To afford costly CNN on resource-constrained mobile/wearable devices, we set to exploit a mobile video stream's \textit{temporal locality}, 
i.e., rich information redundancy among consecutive video frames~\cite{conf/mobisys/MathurLBBFK17,conf/sensys/ChenRDBB15,conf/mobisys/LocLB17}.
Accordingly, a \engine{} can \textit{cache} results when it executes CNN over a mobile video, by using input frame contents as cache keys and inference results as cache values. 
Such caching is expected to reduce the engine's resource demand significantly. 

Towards effective caching and result reusing, we face two major challenges.
\textit{1) Reusable results lookup:} 
Classic caches, e.g., the web browser cache, look up cached values (e.g., web pages) based on key \textit{equivalence} (e.g., identical URLs).
This does not apply to a CNN cache: 
its keys, i.e., mobile video contents, often undergo moderate scene variation over time.
The variation is caused by environmental changes such as user/camera motion, object appearance, and illumination changes~\cite{UCF101}.
A CNN cache must systematically tolerate the variations and evaluate key \textit{similarity}. 
In doing so, the engine must trade off among cacheability, overhead, and model accuracy. 
\textit{2) Fine-grained reuse within a CNN:}
In a CNN model, expensive computations spread across multiple layers.
Besides caching the CNN's final inference outputs, 
the engine should cache the intermediate results (i.e., feature maps) produced by the internal layers. 
Furthermore, the engine should reuse the cached feature maps at fine spatial granularity. 
However, feature maps are high-volume, high-dimensional, barely interpretable data. 
It can be both expensive to inspect them and difficult to assess their similarity.



Few deep learning engines address the two challenges simultaneously.
Commodity engines~\cite{TensorFlow,caffe2,ncnn} process video frames in independent inference tasks with no reuse in between. 
A few recent research prototypes~\cite{conf/mobisys/LocLB17,cavigelli2017cbinfer} incorporate ad-hoc cache designs:
they either look up reusable results based on pixel-wise equivalence of image regions, or perform expensive cache lookup over feature maps at all layers inside a CNN. 
As a result, they often suffer from low cacheability and high lookup overhead, leaving much caching benefit untapped.

To this end, we advocate a principled cache design called \framework. 
The key ideas of \sys{}, as shown in Figure~\ref{fig:overview}, are that i) it discovers reusable image regions by exploiting \textit{the input video's internal structure}, for which it borrows the wisdom from decades of video research~\cite{tham1998novel,zhu1997new,barjatya2004block}; 
ii) it propagates the discovered reusable regions within a CNN by exploiting \textit{the CNN's internal structure}.

As shown in Figure~\ref{fig:overview},
\sys{} stores recent input frames as cache keys and stores recent feature maps for individual CNN layers as cache values. 
To manage the cache, it provides two core mechanisms.
\begin{itemize}
\item 
At the engine input, \sys{} performs cache key lookup: it partitions each video frame into fine-grained regions and searches for similar regions in (cached) recent input frames. 
It does so by running its region matcher. 
Inspired by video compression~\cite{zhu1997new}, the matcher searches neighboring regions in specific patterns guided by video motion heuristics.
\sys{} keeps merging adjacent discovered regions in order to tackle \textit{cache erosion}, i.e., diminishing reusability at deeper layers. 

In contrast to ad-hoc image comparison used by prior CNN caches~\cite{conf/mobisys/LocLB17,cavigelli2017cbinfer}, our matcher is more robust to the aforementioned \variation; 
the matcher runs fast to process more than 1,000 227$\times$227 frames per second.


\item 
Into the CNN execution, \sys{} maps the matched regions on input images to \textit{reusable} regions on feature maps.
It propagates the reusable regions across the feature maps of all CNN layers. 
At each layer, \sys{} transforms the reusable region boundaries based on the operators of this layer; 
it fills the reusable regions with cached feature map values in lieu of actual CNN execution. 
During the process, \sys{} weaves cache queries into CNN computations, keeping the cache queries transparent to CNN models. 

\end{itemize}

With these two mechanisms, \sys{} runs its region matcher only \textit{once} per video frame at the input; it then loads cached feature maps at \textit{all} layers inside CNN. 
This contrasts to ad-hoc approaches that repeat matching processes over both images and feature maps, in and out of CNN.
Our rationale is that, while humans have reliable heuristics on similarity of image contents (which allows \sys{} to assess cache key similarity), they still lack knowledge on evaluating similarity of CNN's internal feature maps that are in disparate dimensions. 
By always treating feature maps as cache values not keys, \sys{} eschews high-cost, low-return searches over them, while still harvesting substantial caching benefit. 

We implement \framework in \emph{ncnn}~\cite{ncnn}, a popular deep learning engine, atop Android 6.0. 
\sys{} executes standard, unmodified CNN models such as ResNet-50~\cite{ResNet}.
\revise{
We evaluate \framework on Nexus 6 with five popular CNN models over two large, real-world video datasets. 
Compared to a baseline engine version without enabling cache, \framework reduces the inference time by 18\% on average and up to 47\%. 
The reduction in inference time by \framework is \textbf{2$\times$} of the reduction achieved by existing CNN caches design~\cite{conf/mobisys/LocLB17}.
\sys{} reduces system energy consumption by around 20\%.
Its incurred accuracy loss is no more than 3\%.
Across all the models, \sys{} uses 2.5 MB -- 44 MB of memory, less than 2\% of the total system DRAM.}

To summarize, we make the following contributions.

\noindent $\bullet$ We present \framework, a principled cache for executing CNN over mobile videos (Section~\ref{sec:overview}).
\sys{} exploits temporal locality in input mobile videos with proven video heuristics (Section~\ref{sec:matching}), propagates cacheable regions across CNN layers with the CNN knowledge (Section~\ref{sec:cache}), and eschews applying video heuristics to CNN internals.

\noindent $\bullet$ We implement \framework in a commodity engine. 
The resultant prototype runs unmodified CNN models, requires zero effort from developers, and is immediately deployable on off-the-shelf Android devices (Section~\ref{sec:impl}).

\noindent $\bullet$ We evaluate \framework on popular CNN models with real-world datasets (Section~\ref{sec:eval}). The results show that \framework can reduce model inference time and energy consumption effectively.

\revise{
The full source code of \framework is at:
\begin{center}
\texttt{https://github.com/xumengwei/DeepCache}
\end{center}
}

\section{Background and Challenges} \label{sec:back}
In this section, we present CNN background and identify the major challenges to cache for continuous mobile vision.

\begin{figure}[t]
	\centering
	\includegraphics[width=0.48\textwidth]{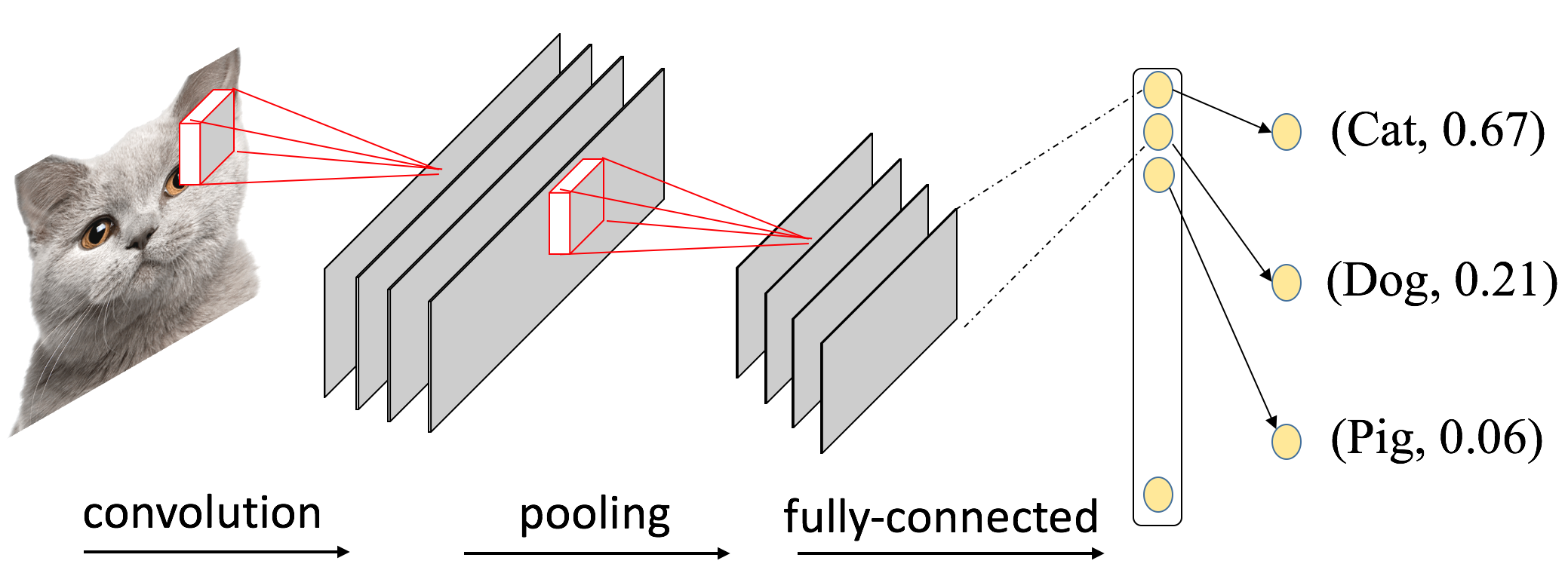}
	\caption{A typical CNN model structure.}
	\label{fig:CNN_Arch}
\end{figure}

\begin{table}[t]
\small
\centering
\begin{tabular}{|l|l|l|l|l|l|l|}
\hline
\textbf{Model} & \textbf{Lib} & \textbf{conv} & \textbf{fc} & \textbf{pl} & \textbf{act} & \textbf{rest} \\ \hline
\multirow{2}{*}{AlexNet~\cite{alexnet}} & TF & 79.2\% & 6.4\% & 11.1\% & 2.7\% & 0.6\% \\ \cline{2-7} 
 & ncnn & 77.9\% & 7.1\% & 12.1\% & 1.8\% & 1.1\% \\ \hline
\multirow{2}{*}{GoogLeNet~\cite{Inception}} & TF & 80.2\% & 0.1\% & 7.5\% & 8.1\% & 4.3\% \\ \cline{2-7} 
 & ncnn & 78.8\% & 0.7\% & 8.6\% & 9.3\% & 2.6\% \\ \hline
\multirow{2}{*}{ResNet-50~\cite{ResNet}} & TF & 91.8\% & 5.8\% & 0.5\% & 1.7\% & 0.2\% \\ \cline{2-7} 
 & ncnn & 93.7\% & 4.9\% & 0.8\% & 0.4\% & 0.2\% \\ \hline
\multirow{2}{*}{YOLO~\cite{yolo}} & TF & 82.4\% & 12.8\% & 2.1\% & 1.8\% & 0.9\%\\ \cline{2-7}
 & ncnn & 84.1\% & 12.2\% & 2.6\% & 0.9\% & 0.2\% \\ \hline
\multirow{2}{*}{Dave-orig~\cite{bojarski2016end}} & TF & 58.8\% & 28.6\% & 4.8\% & 2.9\% & 5.2\% \\ \cline{2-7} 
 & ncnn & 62.7\% & 25.9\% & 5.8\% & 3.7\% & 1.9\% \\ \hline
\end{tabular}
\caption{Processing time breakdown of popular CNN models, showing that convolutional layers dominate the time. 
Layer types: convolutional (conv); fully-connected (fc); pooling (pl), activation (act). 
Hardware: Nexus 6. 
Engines: Tensorflow (TF)~\cite{TensorFlow}; ncnn~\cite{ncnn}.
}
\label{tab:latency_breakdown}
\end{table}

\subsection{Convolutional Neural Network}\label{sec:cnn}
Convolutional Neural Network (CNN) is the state-of-the-art algorithm in many computer vision tasks, and is recently adopted in many mobile scenarios~\cite{conf/mobisys/ZengCZ17,TFCamera,conf/www/YaoHZZA17,conf/ica3pp/OuLSE17,TensorZoom,conf/mobisys/MathurLBBFK17}.
As shown in Figure~\ref{fig:CNN_Arch}, a typical CNN model repeatedly uses convolution and pooling layers to extract features from the whole image, and then applies fully-connected layers (fc) to finalize the 	vision tasks.
Convolutional layers (conv) apply kernels on the input data to extract embedded visual characteristics and generate output data (called \textit{feature map}).
For continuous mobile vision, CNN inference operates only on one single segment of data (i.e., an RGB image) at a time.

\paragraph{Convolutional layers are hotspots}
Among all layer types, convolutional layers are the primary performance hotspots.
\revise{
We summarize the latency breakdown of five popular CNN models in Table~\ref{tab:latency_breakdown}.
We use two libraries that support deep learning inference on Android to run these models on a Nexus 6 device: TensorFlow~\cite{TensorFlow} and ncnn~\cite{ncnn}.
It should be noted that each layer type (e.g., a convolutional layer) can have multiple instances in a model.}
In the breakdown, convolutional layers dominate the processing time, contributing at least 60\% and even up to 90\% (ResNet-50). 
This observation motivates us to focus on caching for convolutional layers in this work.

\subsection{Objective and Challenges}
\label{sec:shrink}
Our overall approach to reduce CNN execution workloads is exploiting temporal locality on a mobile video stream.
That is, consecutive video frames often have substantial similar or overlapped regions.
In general, temporal locality in videos has been known for decades and widely exploited for video compression standards~\cite{le1991mpeg,richardson2004h}.
It is particularly pronounced in mobile videos: mobile devices (e.g., smartphones and glasses), when performing continuous vision tasks~\cite{conf/mobisys/LocLB17,conf/sensys/ChenRDBB15}, capture similar but non-identical image regions continuously. 
To this end, a deep learning engine can cache the CNN execution outcome from processing earlier frames for reuse in processing a later frame.
Of the cache, the \textit{keys} are input image contents and the \textit{values} are the corresponding inference results, i.e., feature maps. 
This objective, while simple, raises a few unique challenges.

\noindent $\bullet$ \textbf{Cache lookup under \variation}~~ 
In general, cache stores key-value pairs. 
Classic caches, e.g., for web browsers or disks, look up cached values (e.g., web pages or disk blocks) by evaluating the \textit{equivalence} of keys (e.g., web URLs or block IDs). 
However, to look up reusable CNN execution results, the cache should evaluate the \textit{similarity} of keys (i.e., input image contents). 
Images consecutively captured in real world can have various aspects of differences for the presence of large variations in camera motion, object appearance, object scale, illumination conditions, etc. Those complicated conditions make it non-trivial to find out \emph{``what should be reused and what should not''}.

\noindent $\bullet$ \textbf{Fine-grained reuse of intermediate results}~~ 
The computation cost of a CNN model spreads over a cascade of internal layers, which produce feature maps as intermediate results. 
An effective CNN cache should store these feature maps and reuse them at fine spatial granularity whenever possible. 
However, deciding reusability for feature maps is challenging: 
since the data volume of feature maps is large, it incurs high overhead for the engine to inspect them; 
since feature maps consist of data points in higher dimension spaces, 
it is difficult for the engine to interpret their semantics. 


\noindent $\bullet$ \textbf{Balancing cacheability, model accuracy, and cache overhead}~~ 
In using cache, the engine will lose CNN model accuracy: it will have to reuse cached values for similar, yet nonidentical, image regions. 
This entails a complex trade-off. 
First, while relaxing the criteria for image similarity boosts cacheability, it also reduces model accuracy.
Second, while more thorough cache lookup improves cacheability, its additional overhead must be justified by sufficient performance gain. 

\begin{figure}[t]
	\centering
	\includegraphics[width=0.45\textwidth]{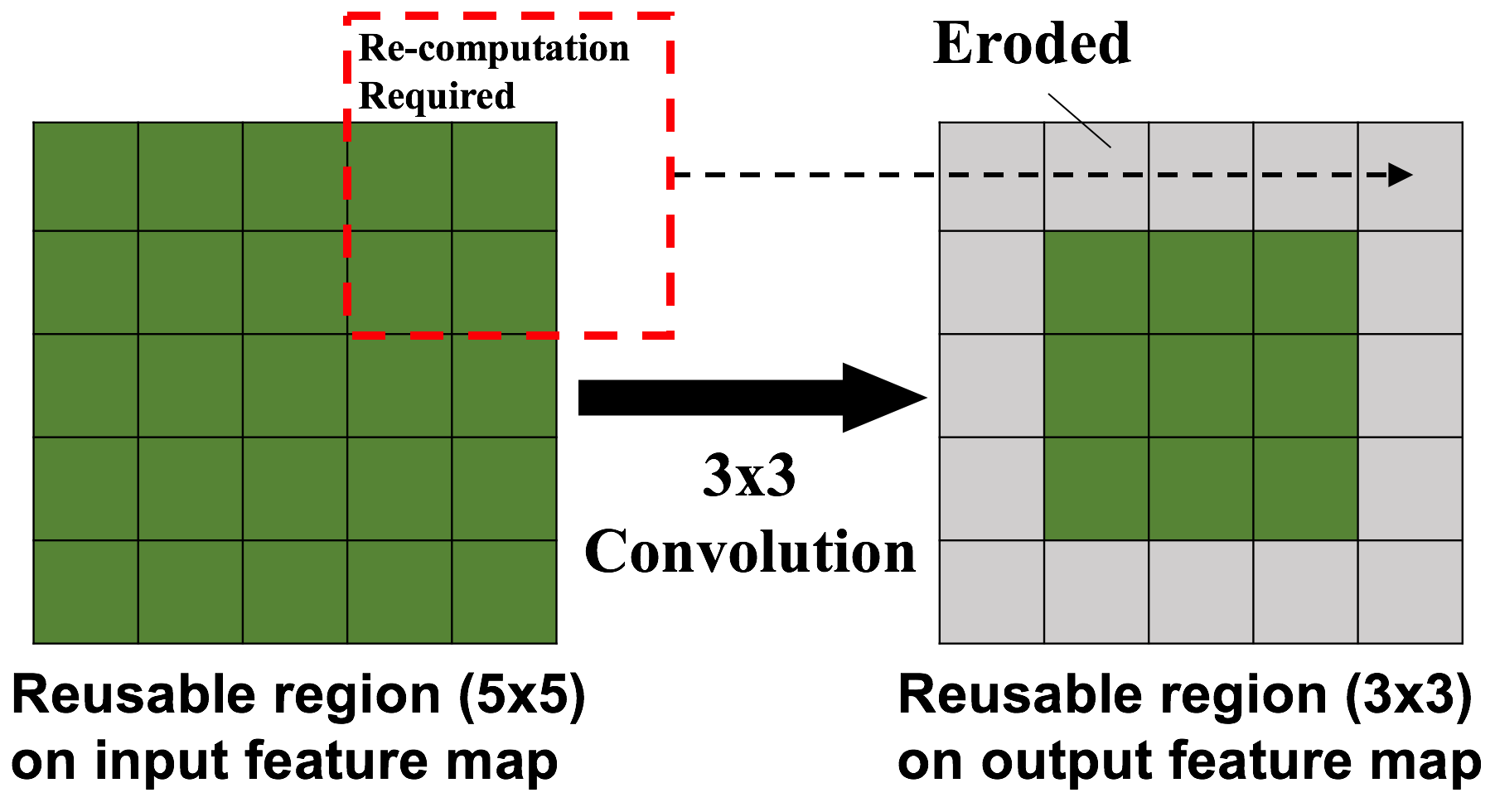}
	\caption{An example showing cache erosion at a convolutional layer (kernel=3x3, stride=1, padding=1).}
	\label{fig:conv_mrect}
\end{figure}

\noindent $\bullet$ \textbf{Battling cache erosion}~~
Of a CNN, reusability tends to diminish at its deeper layers, a behavior we dubbed \textit{cache erosion}. 
More specifically, given an input image region deemed as similar (reusable) to an existing region of previous frame, the amount of reusable results on each layer's feature map shrinks as execution progresses into the model.
Figure~\ref{fig:conv_mrect} shows an example of a convolutional layer, for which the input is a cached region of 5x5 pixels.
However, the peripheral pixels (in gray) in the output cannot be loaded from cache as the central ones (in green), and must be exhaustively computed. 
This is because these peripheral pixels are derived from both reusable and non-reusable results in the input feature map. 
As a result, the reusable region has eroded.

Among various CNN layers, convolution, pooling, and LRN erode cache as above; 
fully-connected layer may completely destroy reusability, since each its output value depends on all its input values, which can hardly be all cached. 
\begin{figure}[t]
	\centering
	\includegraphics[width=0.48\textwidth]{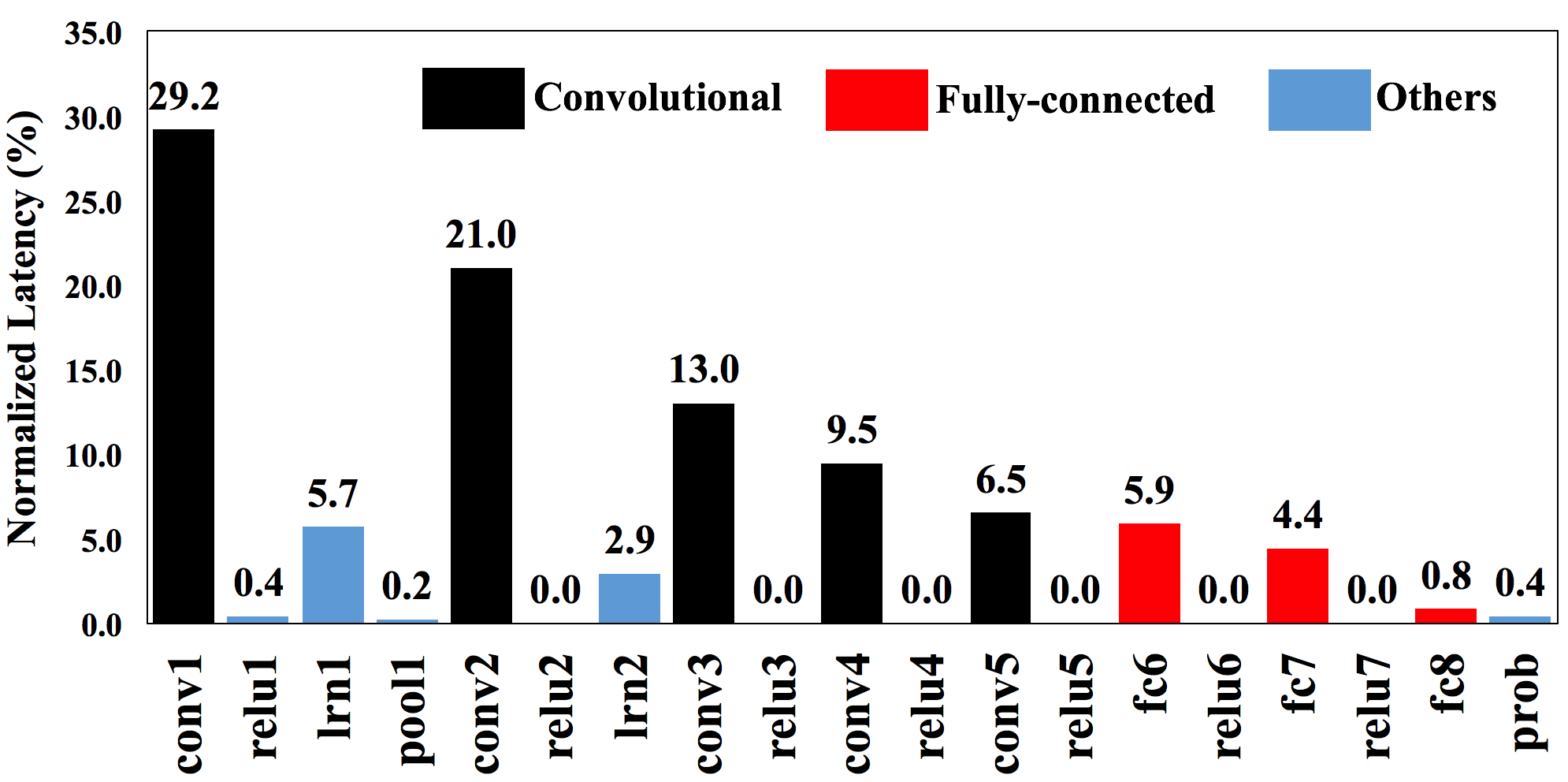}
	\caption{Latency breakdown at layer granularity for AlexNet~\cite{alexnet}. Layers are presented at the order of execution: left-side layer will be executed first and the output will be fed to the right-side layer as input.}
	\label{fig:alexnet}
\end{figure}

Fortunately, in most CNN models, early layers contribute most of the computation cost and also suffer less cache erosion. 
Fully-connected layers come last in a CNN and contribute minor cost.
These are exemplified in Figure~\ref{fig:alexnet}, which breaks down the execution latency of a popular CNN model. 
Of the total latency, only 11.5\% is contributed by fully-connected layers, while the remaining 88.5\% is contributed by earlier layers that can benefit from cache. 
To further tackle cache erosion, we merge reusable regions into the largest possible ones, as will be discussed in Section~\ref{sec:matching}. 

\section{System Overview} 
\label{sec:overview}

\begin{figure*}[t]
	\centering
	\includegraphics[width=0.98\textwidth]{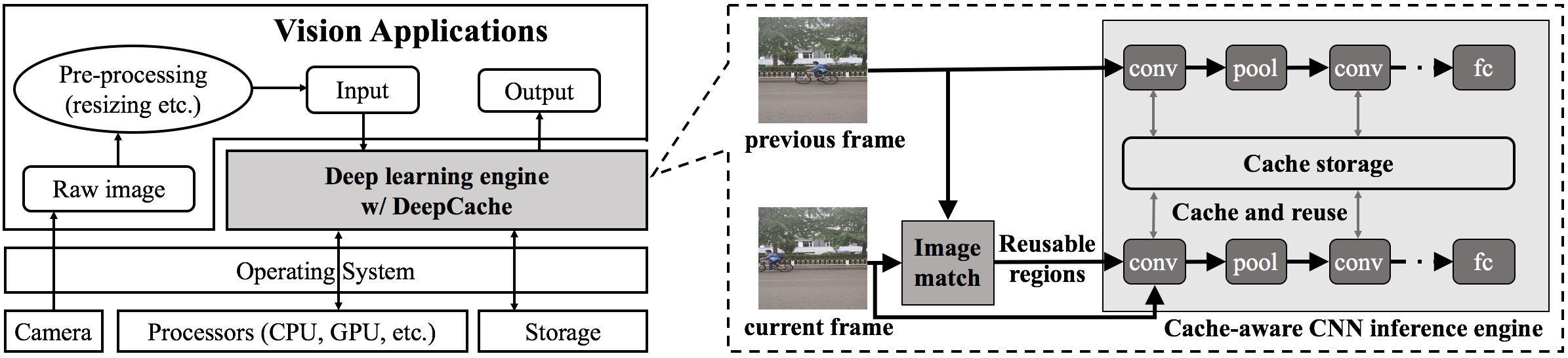}
	\caption{Overall workflow of \framework.
	Black arrows: data flows in CNN execution; green arrows: the reuse of cached data between two consecutive frames.
	}
	\label{fig:workflow}
\end{figure*}
\revise{

\sys{} reduces CNN execution workloads by computation reuse.
The key advantages of \sys{} include: 
1) \textit{\textbf{No cloud offloading}}: \framework completely runs on a mobile/wearable device without any offloading onto the cloud.
2) \textit{\textbf{Widely deployable}}: \sys{} works well with popular CNN models.
3) \textit{\textbf{Transparency and zero developer-effort}}: \sys{} caches inference results for \textit{unmodified} CNN models, without requiring the developers to re-train the models or tuning the parameters.}
This contrasts to disruptive CNN cache designs~\cite{cavigelli2017cbinfer}.
In addition, \sys{} exposes optional APIs for apps to fine-control cache behaviors (Section~\ref{sec:impl}), analogous to that a browser cache exposes various policy knobs to web apps~\cite{webcache}.
4) \textit{Minor accuracy loss}: \sys{} minimizes the model accuracy loss, which it trades for cacheability. 

Figure~\ref{fig:workflow} shows the architecture of \framework.
\sys{}  works as a lightweight extension to a commodity deep learning inference engine.
It augments existing  model inference with cache, while keeping all other engine components unchanged, including loading CNN model file, ingesting video from the camera, pre-processing video frames, executing CNN models on CPU/GPU, and emitting the final output.

\paragraph{\sys{} in a nutshell}
For a CNN model, \sys{} maintains a cache, covering the model's input as well as its internal layers. 
The cache stores recent video frames for the model input, and recent feature maps for the internal layers. 
The \textit{cache keys} are equal-sized, fine-grained regions on the cached input frames. 
The \textit{cache values} are the cached feature maps produced by the layers.

For a new input frame, \sys{} does one-time key lookup by searching for \textit{similar} regions in cached input images.
Upon match, 
\sys{} supplies the engine with corresponding cache values, i.e., feature map regions directly derived from the matched image regions. 
It further propagates these regions to deeper CNN layers:
between layer $L_n$ and layer $L_{n+1}$, \sys{} maps the reusable regions on $L_n$'s feature map to $L_{n+1}$'s feature map. 
It fills these regions with cached feature maps without further key lookup, i.e., search, over these feature maps. 


\paragraph{Key lookup: image region matcher (Section~\ref{sec:matching})}
Prior to executing CNN over a newly captured video frame, \sys{} partitions the frame into equal-sized, fine-grained regions (default 10x10 pixels in each region).  
For each region, \sys{} searches for similar regions in recent input frames. 
It does so based on diamond search~\cite{zhu1997new}, a famous algorithm in motion estimation and video compression.
Compared to ad-hoc, roll-your-own image match, a mature algorithm is not only proven by decades of practice but also may enjoy pervasive hardware acceleration, e.g., hardware video encoders on mobile SoCs~\cite{videoencoder}. 
The matching results are a set of rectangle-to-rectangle mappings, e.g., $(x_i, y_i, w, h) \rightarrow (x'_i, y'_i, w, h)$, where $(x_i, y_i)$ ($(x'_i,y'_i)$) is the left-top point in the current (previous) frame and w (h) is the width (height) of a certain rectangle.

\paragraph{Value mapping: propagating regions across layers (Section~\ref{sec:cache})}
After matching image regions on a new input frame, \sys{} sends the frame and the discovered reusable regions into the CNN model.
\sys{} augments normal CNN execution with three functions. 
First, it propagates the mappings between reusable regions (on the new frame) and cached regions (on an old frame), alongside the input data.
Second, the spatial convolution operation skips computation for the reusable regions and instead directly loads from cached feature maps.
Third, \sys{} caches the output feature map at each convolutional layer 
for future inference. 
\section{Image Block Matching} \label{sec:matching}

\begin{figure}[t]
	\centering
	\includegraphics[width=0.48\textwidth]{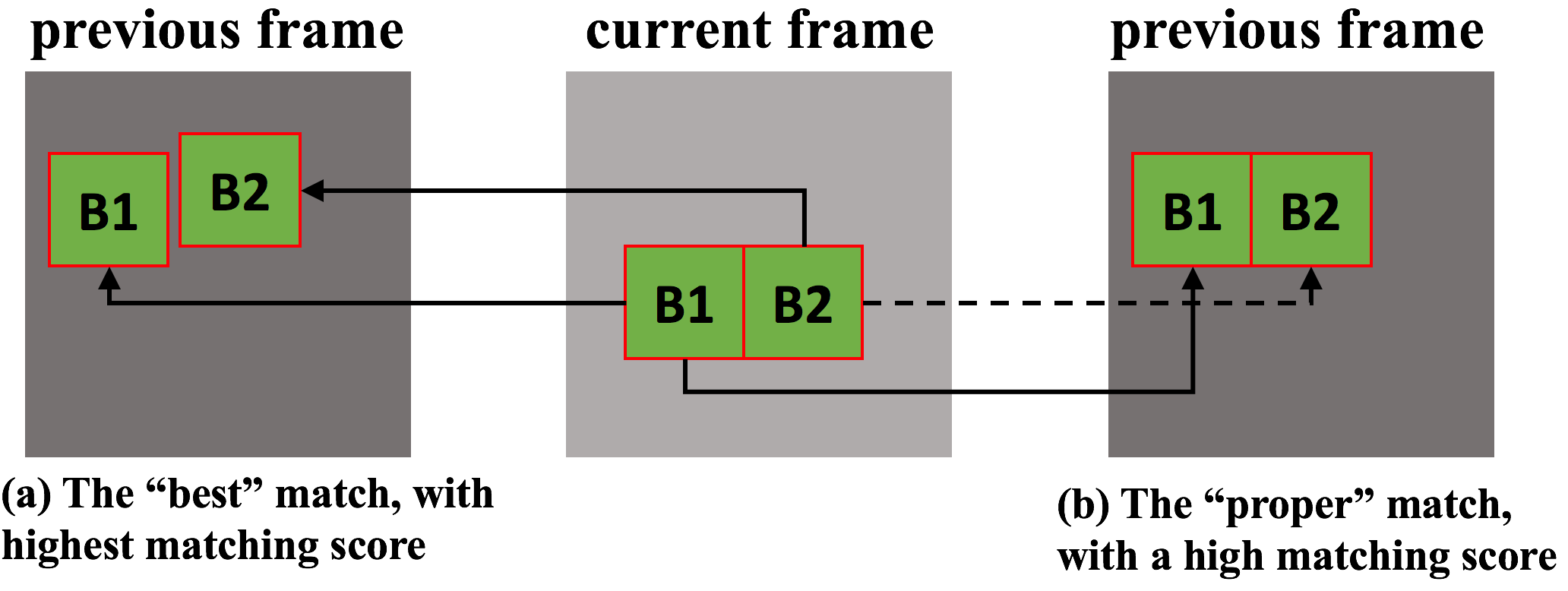}
	\caption{Two matching examples, showing that the best matched block are not always desirable.}
	\label{fig:proper_match}
\end{figure}

Now we present the detailed design of our region matcher and how it deals with \shrink.
The goal of our image matching algorithm is to find ``similar'' regions (rectangles) between two images.
There are two ways to match: \textit{block-wise matching} and \textit{pixel-wise matching}.
Theoretically, identifying each pixel's matching level (pixel-wise matching) and reusing its cached results can be more fine-grained and minimize the model accuracy loss.
However, we have observed that even similar scenes in two sequential images can have relatively low matching scores of corresponding pixels (pixel mutation), due to barely unnoticeable environment variations such as light and moving objects.
Those ``unmatched'' pixels can lead to significant reduction of cache reuse due to the \shrink mentioned in Section~\ref{sec:shrink}.
Thus, we use block-wise matching rather than pixel-wise matching, taking a block (e.g., 10x10 pixels) as the basic unit to tell if it's successfully matched to a corresponding block in the previous image.
In this way, a mutated pixel will not affect the block-wise matching decision if other surrounding pixels in the block are well matched.

Two principles should be considered into the design of our block-wise matching algorithm.
First, the matching algorithm should run fast, keeping the processing overhead negligible compared to the improvement gained via cache reuse.
Second, we want the resulted blocks to be likely merged into larger blocks.
The second principle is exemplified by the case shown in Figure~\ref{fig:proper_match}: match(a) might have the highest matching scores for block B1 and B2, but it's not suitable in our cache mechanism since these small reusable blocks will quickly vanish after several layers due to \shrink (Section~\ref{sec:shrink}).
Imagine that B1 and B2 have size 5x5, and the convolutional kernel is 3x3.
After the \shrink, the reusable regions become two 3x3 rectangles, 18 pixels in total.
\revise{By contrast, match(b) finds two adjacent blocks in current frame that are similar to the blocks in previous frame, so that these two blocks can be merged into a larger one.
In this case, the reusable region becomes one 3x8 rectangle after convolution, 24 pixels in total.}

The overall flow of our matching algorithm is as follows.

\noindent $\bullet$ \textbf{Step 1}. The current frame (image) is divided into an NxN grid, where each grid block contains certain number of pixels.

\noindent $\bullet$ \textbf{Step 2}. For each divided grid block we find the most matched same-size block in previous frame.
Here, we denote the left-top point of \emph{i}-th block (i = 1 to $N^2$) in current frame as $(x_i,y_i)$, and the corresponding matched block position in previous frame as $(x'_i,y'_i)$.
We leverage the \emph{diamond search}~\cite{zhu1997new} algorithm which is widely used in video compression to quickly identify the most matched block.
The matching level (similarity) between two image blocks is represented by the \emph{PSNR}~\cite{zhu1997new} metric: higher \emph{PSNR} indicates that two blocks are more similar.

\noindent $\bullet$ \textbf{Step 3}. We calculate the average block movement $(M_x, M_y)$ as the mean movement of the matched blocks whose \emph{PSNR} is larger than the given threshold $\mathcal{T}$.
\[(M_x,M_y)=(\dfrac{\Sigma{(x'_i-x_i)}}{K},\dfrac{\Sigma{(y'_i-y_i)}}{K}), \langle (x_i,y_i),(x'_i,y'_i) \rangle \in \mathcal{S}\]
where $\mathcal{S}$ is the collection of matched block pair whose PSNR is larger than $\mathcal{T}$, and K is the cardinality of $\mathcal{S}$.

\noindent $\bullet$ \textbf{Step 4}. For each block $(x_i,y_i)$ in the current frame, we calculate its \emph{PSNR} with block $(x_i+M_x,y_i+M_y)$ in the previous frame.
If \emph{PSNR} is larger than $\mathcal{T}$, these two blocks are considered to be properly matched.

\noindent $\bullet$ \textbf{Step 5}. We merge the small blocks that are properly matched in last step to larger ones.
For example, if $(x_i,y_i)$ and $(x_j,y_j)$ in current frame are adjacent, then their matched blocks in Step 4 should also be adjacent since they share the same offset $(M_x,M_y)$.
Thus, we can directly merge them into a larger rectangle as well as their matched blocks.

\begin{figure}[t]
	\centering
	\includegraphics[width=0.45\textwidth]{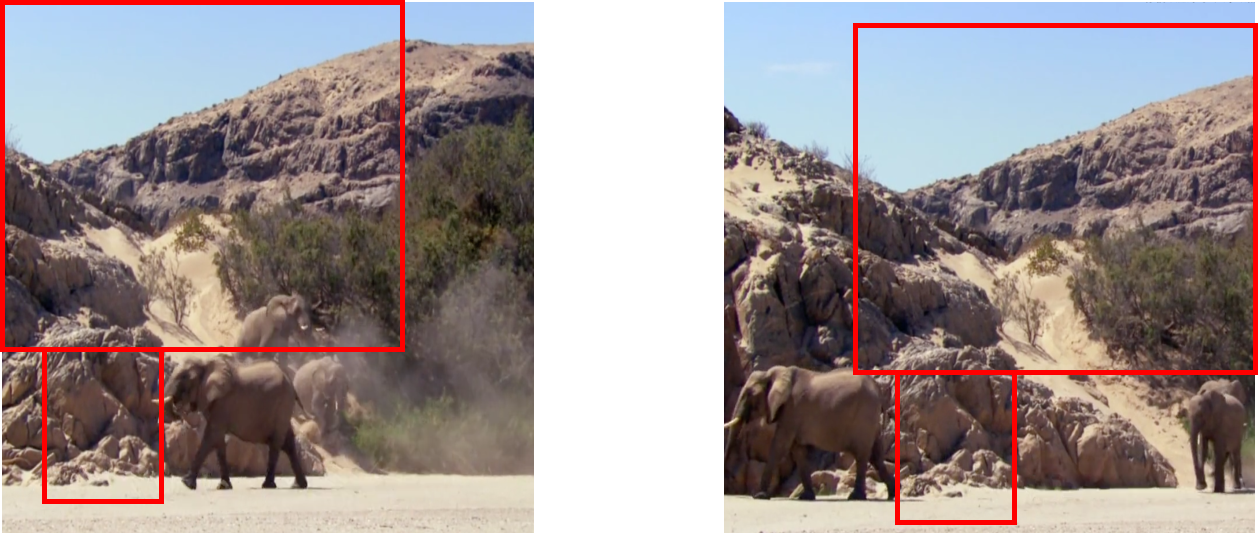}
	\caption{Matched rectangles in two consecutive images via our proposed algorithm.}
	\label{fig:match_elephant}
\end{figure}

Figure~\ref{fig:match_elephant} shows an output example of applying our matching algorithm on two consecutively captured images.
As observed, the second frame image is different from the first one in two aspects.
First, the camera is moving, so the overall background also moves in certain direction.
This movement is captured in Step 3 by looking into the movement of each small block and combining them together.
Second, the objects in sight are also moving.
Those moved objects (regions) should be detected and marked as non-reusable.
This detection is achieved in Step 4.

Our experiments show that most of the processing time of the above matching algorithm is spent at Step 2 and Step 4.
In Step 2, we need to explore the previous frame to identify the most matched block for every block in current image.
We can accelerate this step by skipping some blocks in current frame, e.g., only matching blocks at $(i*k)$-th row and $(j*k)$-th column ($i*k,j*k\leqslant N$).
Theoretically, a 2-skip (\emph{k}=2) can save 75\% of the computation time in this step, and a higher k can even achieve better improvements.
However, a higher \emph{k} might also result in inappropriately calculated $(M_x,M_y)$, resulting in fewer blocks to be properly matched at the last step.
We can further accelerate the computation of Step 4 by reusing the results in Step 2 since both of them need to calculate \emph{PSNR} between two blocks.
More specifically, if the \emph{PSNR} between $(x_i,y_i)$ (current frame) and $(x_i+M_x,y_i+M_y)$ (previous frame) is already calculated in Step 2, we simply reuse the result.
We demonstrate the efficiency of our proposed algorithm as well as these acceleration approaches in Section~\ref{sec:eval_matching}.
\section{Cache Mechanisms Inside Model Execution} 

\label{sec:cache}
\begin{table}[t]
\scriptsize
\centering
\begin{tabular}{|l|L{1.7cm}|L{4.2cm}|} \hline
	\textbf{Layer Type} & \textbf{Layer Parameters} & \textbf{Output}($\mathcal{D}_t$)\\ \hline
	Convolution & kernel=k x k & \multirow{2}{*}{\shortstack[l]{
	$x'=\ceil{(x+p)/s},y'=\ceil{(y+p)/s}$\\$w'=\floor{(w-k)/s},h'=\floor{(h-k)/s}$}}\\ \cline{1-1}
	Pooling & stride=s, padding=p & \\ \hline
	LRN~\cite{layer-lrn} & radius=r & \shortstack[l]{$x'=x+r,y'=y+r$\\$w'=w-2*r,h'=h-2*r$}\\ \hline
	Concat~\cite{layer-concat} & input number=\emph{N} & overlapped region of these \emph{N} rectangles\\ \hline
	Fully-connected & \multirow{2}{*}{/} & \multirow{2}{*}{$(x',y',w',h') = (0, 0, 0, 0)$}\\ \cline{1-1}
	Softmax~\cite{layer-softmax} & & \\ \hline
	Others & / & $(x',y',w',h') = (x, y, w, h)$\\ \hline
\end{tabular}
\caption{Transformation of reusable region boundaries for layer type ($\mathcal{D}_t$). Input region is a rectangle $(x,y,w,h)$.}
\label{tab:mrect}
\end{table}
To cache a model's internal inference results, 
\sys{} provides two facilities:
propagation and reuse.


\paragraph{Propagation}
To reuse the computation results inside CNN inference, \framework needs to identify which regions can be reused and where they are mapped to for each layer's output.
As previously explained in Section~\ref{sec:shrink}, the mappings obtained by matching raw images (Section~\ref{sec:matching}) need to be dynamically adjusted at inference runtime.
This adjusting should also be performed on the corresponding cached blocks of previous frame.
Obviously, the strategy about how reusable regions are adjusted is based on the forward operation of different layer types.
More specifically, a caching mapping $(x_i,y_i,w,h) \rightarrow (x'_i,y'_i,w,h)$ will be adjusted to $\mathcal{D_\emph{t}}(x_i,y_i,w,h) \rightarrow \mathcal{D_\emph{t}}(x'_i,y'_i,w,h)$ after operation \emph{t}, where function $\mathcal{D}_t$ indicates how a reusable region should be adjusted after going through a layer type \emph{t}.
We show the design details of $\mathcal{D}_t$ for every layer type in Table~\ref{tab:mrect}.
There are three main types of layers in consideration of how they affect the reusable regions:
1) Locally coherent layers that computes each pixel based on a part of input such as convolutional and pooling layers. These layers will diminish the reusable regions.
2) Fully-connected and softmax layers that connect each neuron in input and output. These layers totally destroy the data localization so that there will be no reusable regions.
3) Activation layers such as ReLu, Sigmoid that produce each output neuron based on the corresponding input neuron. These layers have no effect on the reusable regions.

\begin{figure}[t]
	\centering
	\includegraphics[width=0.48\textwidth]{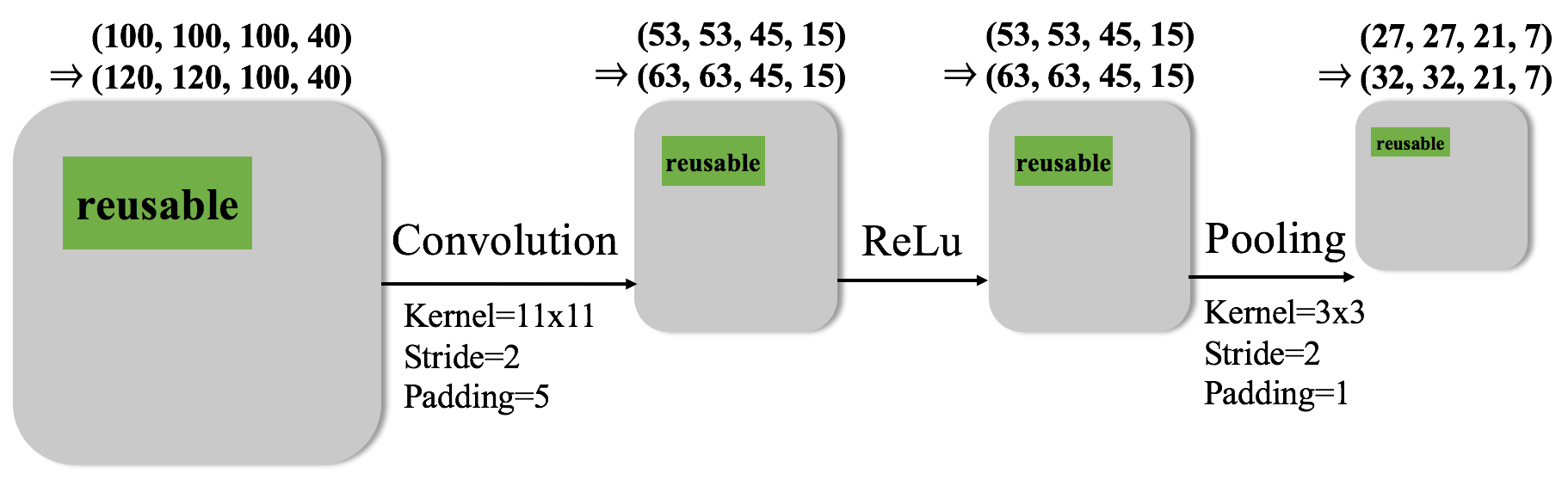}
	\caption{An example showing how the mappings obtained by image matching are adjusted during the CNN inference.
	The grey rectangles represent the data flow among CNN layers, while the black arrows represent the operations performed on the data.
	}
	\label{fig:mrect}
\end{figure}

Figure~\ref{fig:mrect} shows an illustrating example about how a reusable region is propagated among different layers.
The current image has been matched to previous image, and a block $(100, 100, 100, 40)$ (left-top=$\langle100, 100\rangle$, width=100, height=40) is identified to be similar to the block $(120, 120, 100, 40)$ of last frame.
This image is the input of a convolutional layer, with kernel=11x11, stride=2, and padding=5.
The reusable region of computational output of this layer can be calculated as $(53, 53, 45, 15)$.
This output is passed to an activation layer (ReLu) as input, but the reusable region is not changed since the activation layer performs just a certain activation function on every single input unit.
Then, the output of ReLu is consumed by a pooling layer, with kernel=3x3, stride=2, padding=1.
Similar to the convolutional layer, the reusable region becomes smaller due to the kernel padding.

\paragraph{Reuse}
After knowing which regions can be reused, \framework customizes the convolutional forward so that the computations of these reusable regions are skipped.
Instead, they will be directly copied from the corresponding cached region from previous frame.
When customizing convolution operations, it's important to achieve good data locality since data movement is one of the computational bottlenecks~\cite{conf/isca/ChenES16} during convolution processing.
To this end, \framework splits the convolution operation into three steps.
First, reusable regions are directly copied from cache of last frame.
Second, a boolean (bit) map is created to specify whether a pixel $(x,y)$ is already cached.
Third, kernel travels on the input feature map and performs convolution only on the non-reusable pixels but skips reusable ones.

\framework caches and reuses the computational results only in convolutional layers for two reasons.
1) As mentioned in Section~\ref{sec:back}, convolution is usually the dominant layer in CNN inference time (e.g., 79.2\% for AlexNet).
2) Caching the intermediate output for other layer types (e.g., pooling) requires additional memory overhead.
In other words, \framework supports caching reuse only in convolutional layers to make proper trade-off among latency improvement and memory overhead.
But it's worth mentioning that we can easily extend our cache mechanism to other layer types.

Though \framework reuses only the computation of similar image blocks, there is still accuracy loss since the matched blocks may not be numerically identical.
For two consecutive frames, the output disparity can be negligible.
However, if the caching goes on for more frames, the accuracy loss might be non-ignorable.
To mitigate the superposition of accuracy loss caused by caching, \framework periodically cleans its cache and calculates a whole new frame every $\mathcal{N}$ frames (default to 10).

\begin{figure}[t]
	\centering
	\includegraphics[width=0.45\textwidth]{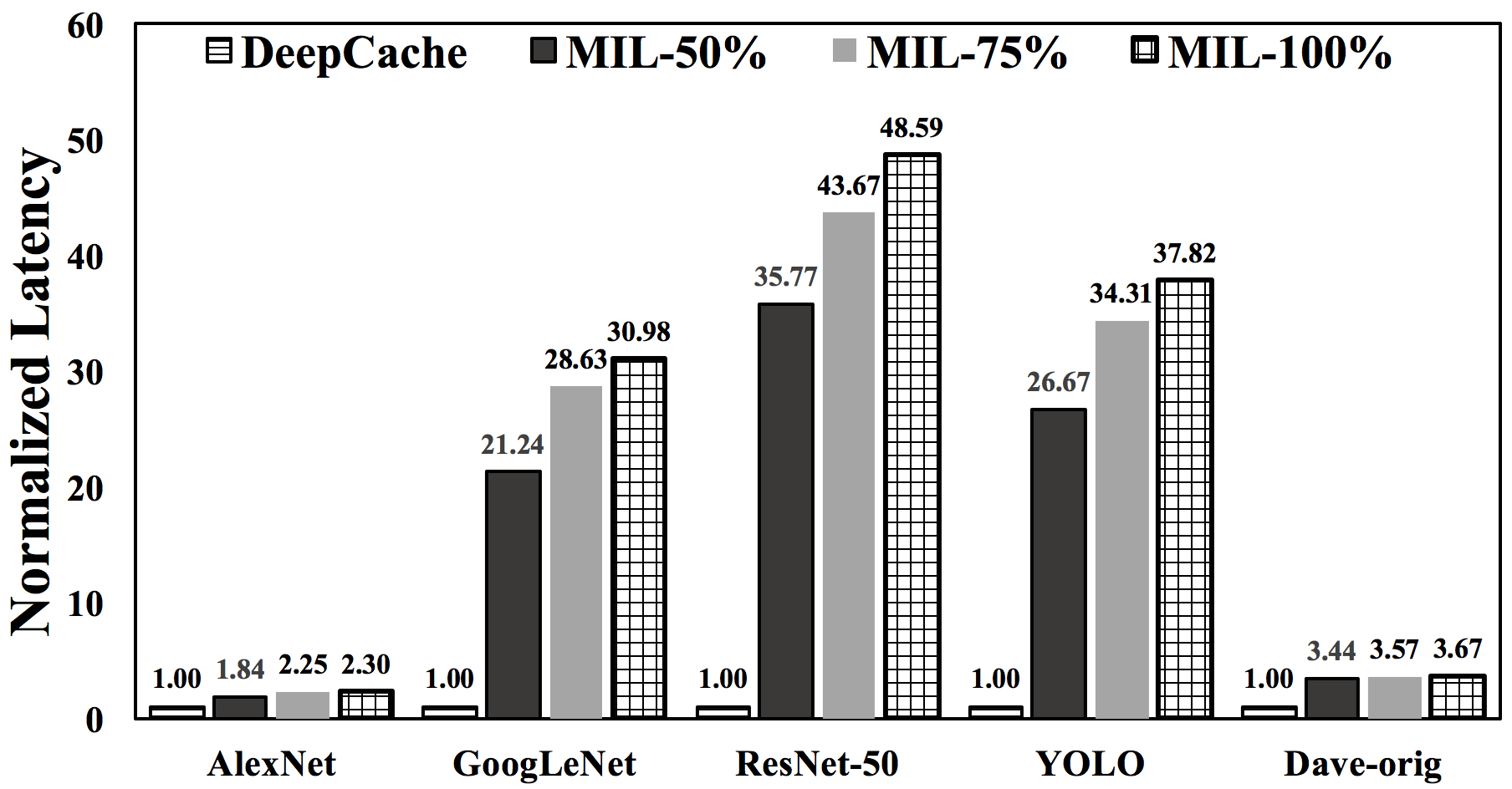}
	\caption{Comparison of match cost (i.e., cache lookup) between \sys{} and ``matching internal layers'' (MIL), an alternative that attempts to match on all internal feature maps. 
	MIL-N\%: matching feature maps at the first N\% convolutional layers in the model.
	Compared to \sys{}, the overhead of MIL is prohibitive.
	}
	\label{fig:eml}
\end{figure}

\paragraph{Compared to matching internal layers}
Cache erosion hurts reusability (Section~\ref{sec:shrink}). 
An ad-hoc approach to mitigating cache erosion would be aggressively \textit{searching} for reusable regions on feature maps~\cite{cavigelli2017cbinfer} cached for all layers, as we call ``matching internal layers'' (abbreviated as MIL).
Hence, this approach not only matches regions on input frames as \sys{} does, but also matches regions on feature maps that are generated during inference. 
By doing so, it essentially treats feature map regions as cache keys and looks them up in cache. 

Conceptually, MIL may help reusability. Yet, we deem it impractical for the following reasons. 

1) \textit{High cost.} 
Cached feature maps are in high volume. 
Scanning them for each input frame is expensive.
Figure~\ref{fig:eml} compares the latency in match (i.e., cache lookup) with \sys{}'s approach (propagation of regions) with that of MIL.
We thoroughly test MIL by varying the number of convolutional layers it attempts to match on. 
The results show that MIL incurs much higher latency than \sys{}, even when MIL only covers 50\% of the total convolutional layers.  
This performance gap can be as large as 35$\times$! (e.g., for ResNet)

2) \textit{Low return.}
Decades of image research have yielded reliable heuristics on image similarity estimation~\cite{zhu1997new,tham1998novel}. 
By contrast, we know much less about evaluating similarity among 
CNN feature maps. 
Hence, when feature maps are used as keys, evaluating their similarity for reuse is fundamentally difficult. 
\revise{One might, for example, devise numerical thresholds for feature map differences~\cite{cavigelli2017cbinfer}. 
However, our experiences suggest this as intractable: good thresholds, if exist at all, are specific to models, layers, or even inputs.
In other words, MIL inevitably requires extra efforts from application/model developers to identify a good threshold for every single layer of a given CNN model.
In comparison, our design of key lookup doesn't need such efforts from developers.}

\section{Implementation} \label{sec:impl}

We implement our image matcher (Section~\ref{sec:matching}) in RenderScript~\cite{RenderScript}, the Android's counterpart of CUDA. 
Thanks to RenderScript, the image matcher execution can be offloaded to GPU for acceleration. 
Since RenderScript is a generic Android API, our image matcher is portable across Android devices. 

We prototype the engine feature of \sys{} atop \emph{ncnn}~\cite{ncnn}, an open-source deep learning engine optimized for mobile (Android and iOS).
\emph{ncnn} works with standard CNN models.
\framework are directly compatible with those models without requiring extra model changes.

For each supported layer type, \emph{ncnn} provides a function \texttt{forward(top\_blob, bottom\_blob)}, where \texttt{top\_blob} and \texttt{bottom\_blob} encapsulate the output and input of this forward step, respectively.
We replace \texttt{forward()} with our customized \texttt{c\_forward(top\_blob, bottom\_blob, c\_blob, c\_regions)}, where \texttt{c\_blob} stores the computation results of current layer from the last frame, and \texttt{c\_regions} specifies which parts can be reused.
\texttt{c\_forward} calculates the output just as \texttt{forward} does, except that \texttt{c\_forward} skips the calculation of cached regions but copies from \texttt{c\_blob} directly.
Before \texttt{c\_forward} invoked, \texttt{c\_regions} will be propagated from last layer.
As mentioned in Section~\ref{sec:cache}, cached regions will erode (conv, pooling) or vanish  (full-connected) during the inference process, thus we use another function named \texttt{reg\_forward} which calculates how cached regions are propagated among different layers.
We also implement some custom layers such as \texttt{atan} that are unsupported in the current \emph{ncnn} but necessary for our benchmark models. 
Overall, our new implementation contains 4,030 lines of code. 


\sys{} is fully compatible with \textit{ncnn} APIs. Any existing vision applications built on \textit{ncnn} will work with \sys{} out of box. 
In addition, \sys{} exposes a few cache parameters, e.g., threshold $\mathcal{T}$ (Section~\ref{sec:matching}), block size N (Section~\ref{sec:matching}), and cache expiration time (Section~\ref{sec:cache}) for developers to optionally fine control \framework behavior.
This is analogous to a browser cache exposing various policy knobs to web apps~\cite{webcache}.

\section{Evaluation} \label{sec:eval}
We thoroughly evaluate \framework using five typical CNN models on two real-world, large-scale datasets.
In summary, \framework saves the execution time of CNN models by 18.2\% on average and up to 47.1\%, while incurring accuracy loss as low as 3\%.
\revise{
In addition, we directly compare \sys{} with the cache mechanism presented in DeepMon~\cite{conf/mobisys/LocLB17}, a cutting-edge deep learning engine, and the results show that \sys{} outperforms DeepMon on all models and all datasets.}


\begin{table*}[!t]
\small
\centering
\begin{tabular}{|l|l|l|c|l|l|} \hline
	\textbf{Application} & \textbf{Model Name} & \textbf{Model Architecture} & \textbf{\# of Conv} & \textbf{Model Output} & \textbf{Dataset}\\ \hline
	\multirow{3}{*}{Activity recognition} & REC\_1 & AlexNet~\cite{alexnet} & 5 & \multirow{3}{*}{human activity type} & \multirow{4}{*}{UCF101~\cite{UCF101}}\\ \cline{2-4}
	& REC\_2 & GoogLeNet~\cite{Inception} & 57 & &\\ \cline{2-4}
	& REC\_3 & ResNet-50~\cite{ResNet} & 53 & &\\ \cline{1-5}
	Object detection & DET & YOLO~\cite{yolo} & 8 & object types and positions &\\ \hline
	Self-driving & DRV & Dave-orig~\cite{bojarski2016end,Autopilot} & 5 & steering angle & Nvidia driving dataset~\cite{nvidia-driving-dataset}\\ \hline
\end{tabular}
\caption{List of CNN models used to evaluate \framework.}
\label{tab:models}
\end{table*}

\begin{figure}[t]
	\centering
	\includegraphics[width=0.45\textwidth]{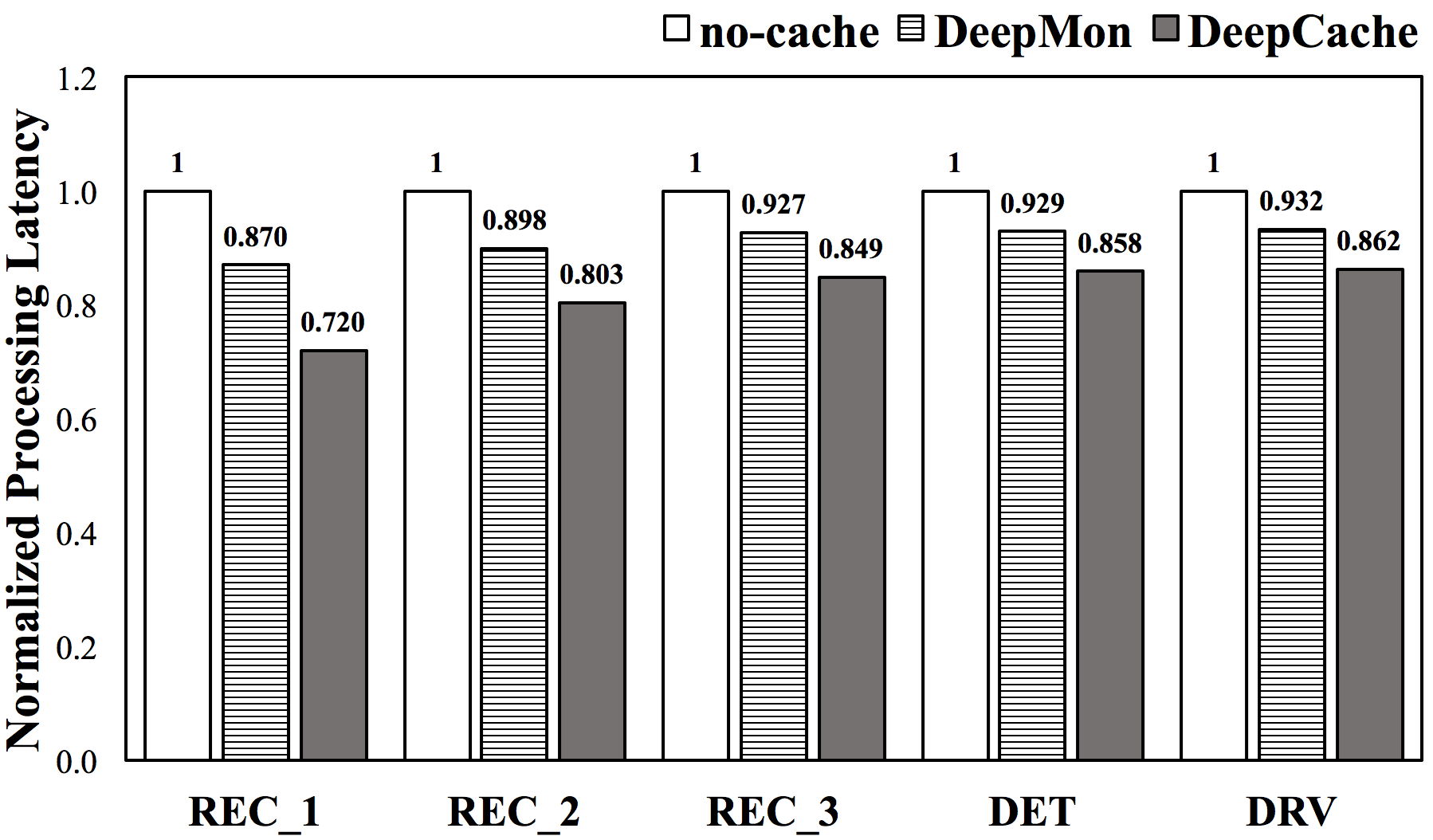}
	\caption{Average processing time of all five CNN models over their test scenarios.}
	\label{fig:latency_overall}
\end{figure}

\begin{figure*}[t]
\centering
\scriptsize
\subfloat[REC\_1 model]{\includegraphics[width=0.5\textwidth]{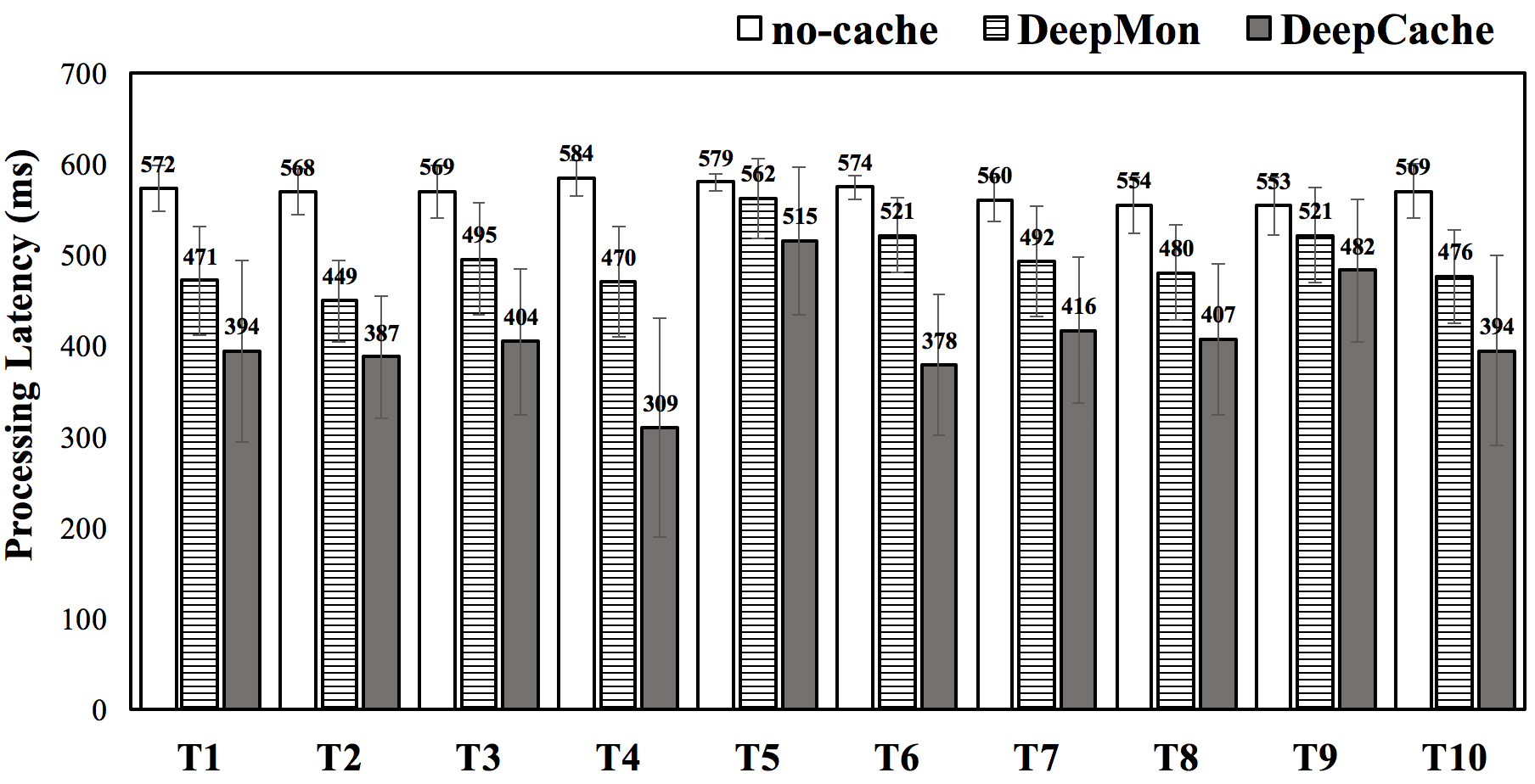}}
\subfloat[REC\_2 model]{\includegraphics[width=0.5\textwidth]{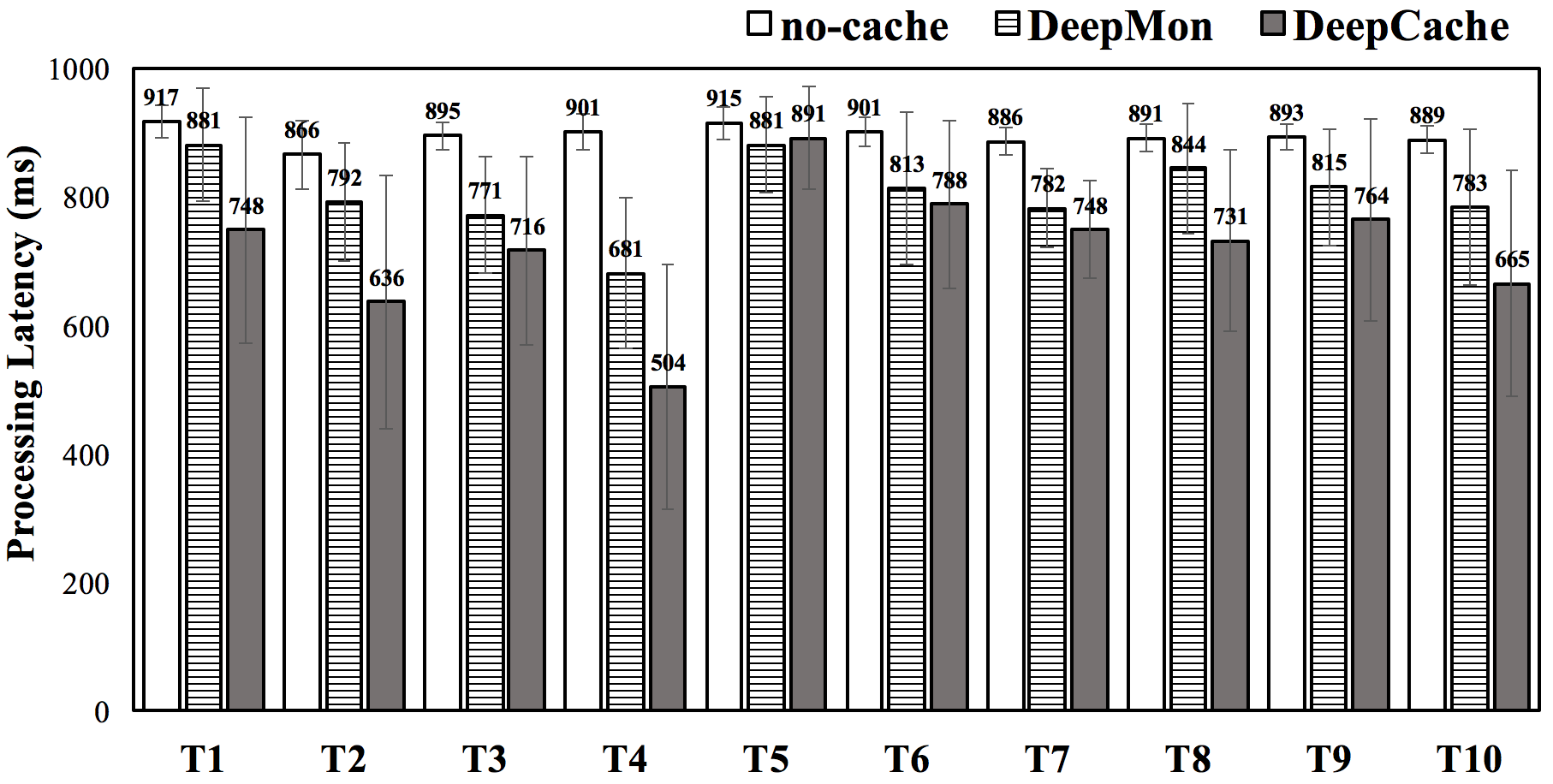}}\\
\subfloat[REC\_3 model]{\includegraphics[width=0.5\textwidth]{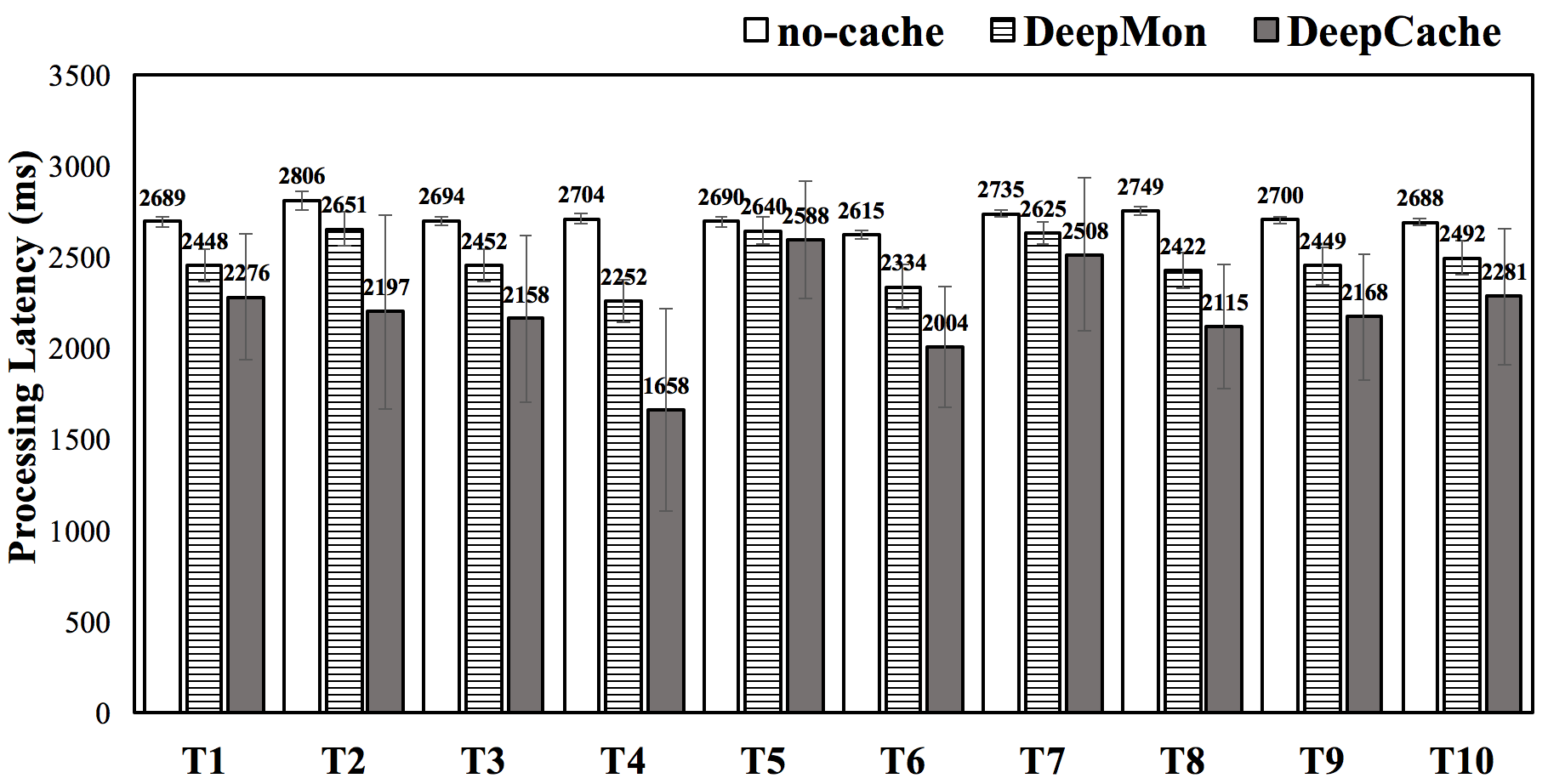}}
\subfloat[DET model]{\includegraphics[width=0.5\textwidth]{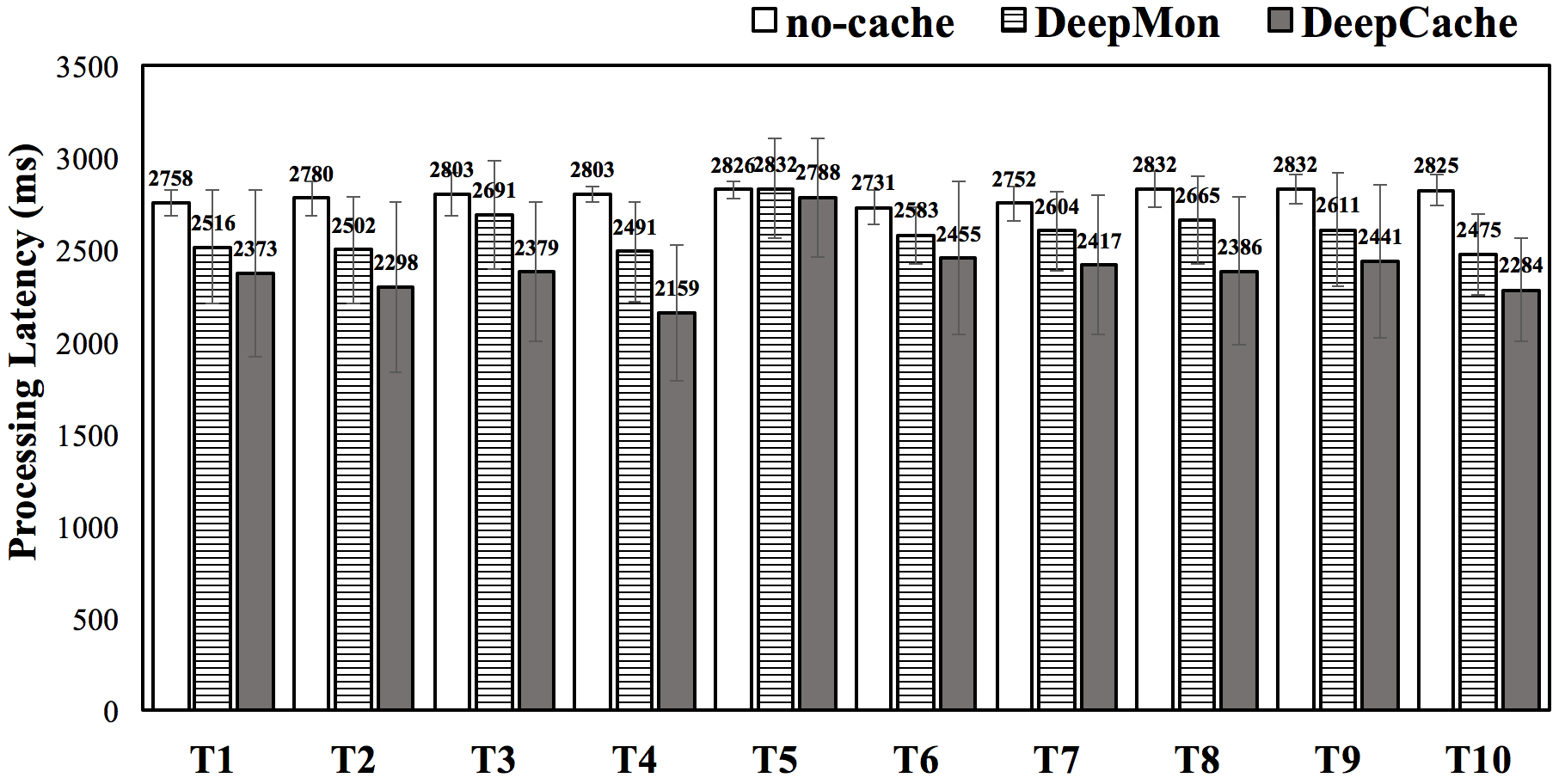}}
\caption{Per-scenario processing time of four CNN models. (For each model, the average time across all scenarios is shown in Figure~\ref{fig:latency_overall}.)}
\label{fig:eval_latency}
\end{figure*}

\subsection{Experimental Setup}
\paragraph{Test Platform}
We use Nexus 6 (Qualcomm 2.7 GHz quad-core CPU; Adreno 420 GPU) with Android 6.0 as the test platform.

\paragraph{Benchmark Datasets}
We use two kinds of datasets to evaluate our framework.
\textbf{UCF101 dataset}~\cite{UCF101} contains 101 types of human activities and 13,421 short videos ($<$ one minute) created for activity recognition.
We randomly select 10 types from these activities and evaluate \framework across them:
\emph{Basketball} (\textbf{T1}), \emph{ApplyEyeMakeup} (\textbf{T2}), \emph{CleanAndJerk} (\textbf{T3}), \emph{Billiards} (\textbf{T4}), \emph{BandMarching} (\textbf{T5}), \emph{ApplyLipstick} (\textbf{T6}), \emph{CliffDiving} (\textbf{T7}), \emph{BrushingTeeth} (\textbf{T8}), \emph{BlowDryHair} (\textbf{T9}), and \emph{BalanceBeam} (\textbf{T10}).
In total, \textbf{55,680} images have been processed in our evaluation for each selected CNN model.
\textbf{Nvidia driving dataset}~\cite{bojarski2016end} is collected by driving on a wide variety of roads and in a diverse set of lighting and weather conditions.
It contains 45,568 static images captured at 10 FPS and the corresponding steering angles made by the driver.
We randomly select 10 scenarios\footnote{A scenario, in video dataset lingo, refers to a video recorded at specific scene, location, and time.} (100 images for each) as the testing set.
\revise{We use ffmpeg~\cite{ffmpeg} tool to extract raw images from the above video datasets and feed the images to \framework sequentially, mimicking video ingestion in real-world continuous vision applications.}


\paragraph{Workloads}
We use a variety of five CNN models to verify \framework as shown in Table~\ref{tab:models}.
For activity recognition, our models (REC\_1, REC\_2, and REC\_3) are pre-trained on ILSVRC 2012 dataset~\cite{imagenet}, and then transferred learned on UCF101.
The architectures of those models are initially used for image classification.
In our case, we use them to run each single image in the video and average the final result~\cite{conf/cvpr/KarpathyTSLSF14}.
For object detection, the model (DET) is trained via Pascal VOC 2007 dataset~\cite{voc}.
The model (DRV) used for self-driving is trained and tested on the Nvidia driving dataset mentioned above.
It is worth mentioning that these CNN models are quite generalized and can be used in many different tasks with few customization efforts.

\paragraph{Metrics}
We use accuracy, processing latency, and power consumption to evaluate the performance of our framework.
To report the \textbf{accuracy} results, we use different metrics to fit into different applications.
We report the top-k accuracy for our activity recognition models, and MSE (Mean Squared Error) as the accuracy for object detection and self-driving tasks because their outputs are continuous values.
Since the dataset used (UCF101) has no labels for object detection, we treat the output of exhaustively running complete model without cache mechanism as ground truth (observed values).
For \textbf{latency}, we log the starting time when \framework receives the image and the ending time when \framework outputs the inference result.
The subtracted duration is reported as the processing latency, including the time spent on image matching and CNN inference.
Finally, we measure the \textbf{energy consumption} via Qualcomm Snapdragon Profiler~\cite{SnapdragonProfiler}.
The baseline of phone's idle state is always subtracted.

\paragraph{\framework Configuration}
If not otherwise specified, we use a default block size of 10x10, the matching threshold $\mathcal{T}$ of 20 in our image matching algorithm (Section~\ref{sec:matching}), and the expiration time N of cache is set as 10 (Section~\ref{sec:overview}).

\paragraph{Comparison to Alternatives}
We experimentally compare the performance of \framework to two alternative approaches: \emph{no-cache}: exhaustively running the complete model without cache reuse (ground truth used in measuring accuracy); 
\emph{DeepMon}~\cite{conf/mobisys/LocLB17}:
the cache mechanism in a state-of-the-art deep learning engine.
To make the comparison fair, we have carefully ported \emph{DeepMon}'s cache to the ncnn engine executed on the CPU of our test platform, where \sys{} also runs. 
Note that we have contrasted the design of \emph{DeepMon} cache with \sys{} (Section~\ref{sec:intro}), and will present more details in related work discussion (Section~\ref{sec:related}).

\subsection{Latency Improvement}
Figure~\ref{fig:latency_overall} summarizes the achieved improvements via applying cache mechanism on average.
Our primary observation is that applying \framework can have substantial latency reduction compared to \emph{no-cache}, i.e., \textbf{18.2\%} on average, while \emph{DeepMon} has only \textbf{8.9\%}.
This improvement varies across different CNN models.
For a relatively small model REC\_1 (5 convolutions, 25 layers in total), \framework results in \textbf{28.1\%} saving up of total processing time on average, while \emph{DeepMon} only has \textbf{13.1\%} improvement.
For a deeper model REC\_2 (57 convolutions, 153 layers in total), the benefit from \framework reduces to \textbf{19.7\%}, while \emph{DeepMon} has only \textbf{10.2\%}.
For DET, \framework can have only \textbf{14.2\%} latency improvement.
The reason is that, differently from other classification models, DET is applied in object detection applications and outputs location-sensitive information.
Thus, many computation-intensive fully-connected layers reside at the end of DET, making the benefit from convolution-layer cache smaller.
The similar situation also applies for the DRV model.

We further illustrate the results under different video benchmarks (UCF101) in Figure~\ref{fig:eval_latency}.
We observe that the performance of \framework can differ a lot under different benchmarks.
Taking REC\_1 as an instance, \framework saves up \textbf{47.1\%} processing time under \emph{Billiards} (\textbf{T4}) scenarios.
We manually check the dataset videos and identify the reasons of such high speedup as following: 1) camera is in slow motion, 2) most objects are still except the player and balls, 3) indoor lighting is stable.
In comparison, \framework has only \textbf{11.0\%} latency improvement when processing \emph{BandMarching} (\textbf{T5}) videos because the camera and most objects (people) in view are moving brokenly.
Similarly, for REC\_3, \framework saves 38.7\% processing time when dealing with \textbf{T4} but only 3.8\% under \textbf{T5}.
Importantly, we observe that \framework consistently beats \emph{DeepMon} for each scenario.

\begin{figure}[t]
	\centering
	\includegraphics[width=0.48\textwidth]{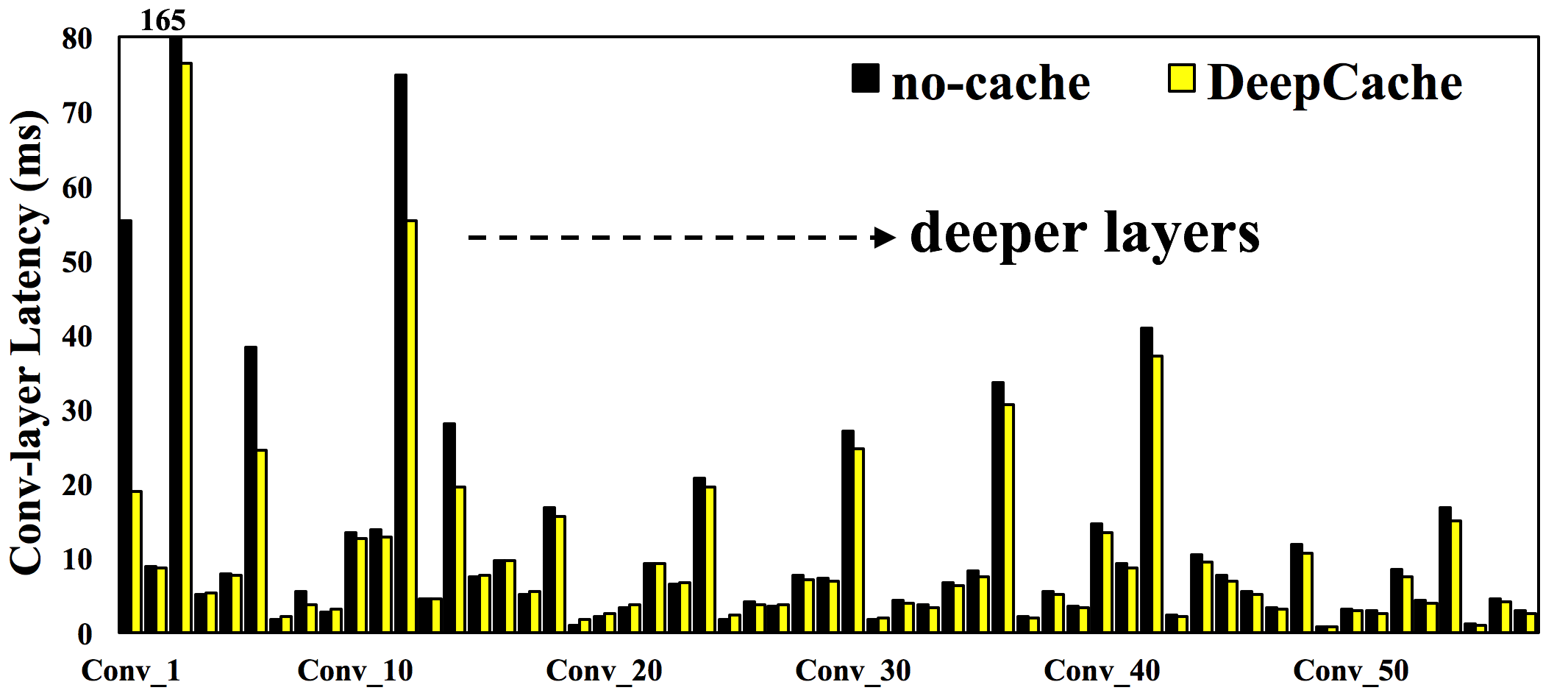}
	\caption{Processing time for individual convolutional layers in model REC\_2.}
	\label{fig:eval_latency_breakdown}
\end{figure}

We further dig into the achieved improvement at each individual convolutional layer.
As shown in Figure~\ref{fig:eval_latency_breakdown}, the latency improvement mainly comes from the first few layers due to the \shrink mentioned previously.
Fortunately, these layers often contribute to the majority of overall latency, indicating that the benefit remains meaningful when models grow deeper.
For example, the third convolutional layer takes 165ms to run, which contributes around 18.4\% to the total model.
\framework is able to save 90.2ms from this single layer since this layer resides at the beginning of the overall model.

\subsection{Accuracy Loss}
\begin{figure}[t]
	\centering
	\includegraphics[width=0.44\textwidth]{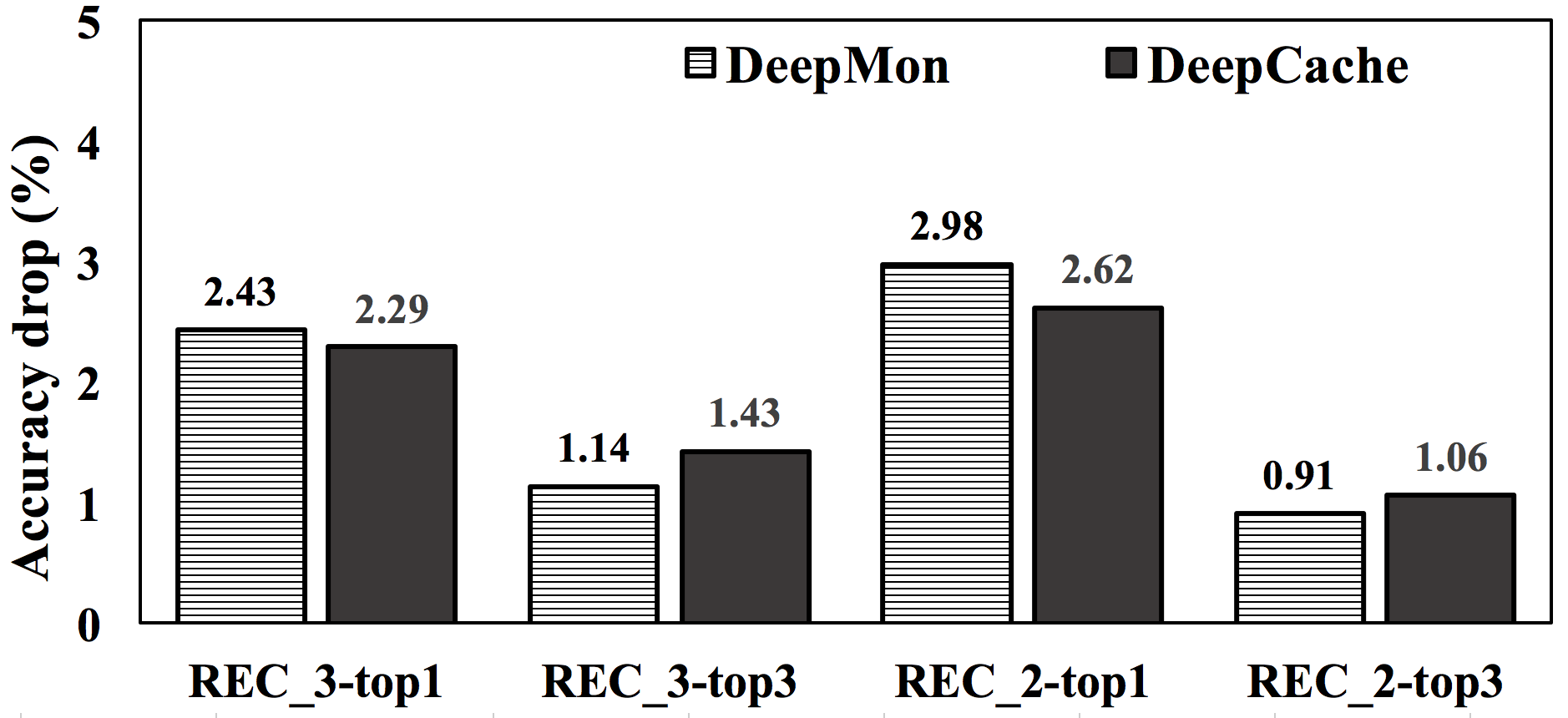}
	\caption{Top-k accuracy drop of \framework.}
	\label{fig:eval_top_accuracy}
\end{figure}
We then investigate how much accuracy \framework compromises in return for the latency benefits.
The top-k accuracy drop for our activity recognition is shown in Figure~\ref{fig:eval_top_accuracy}.
In overall, \framework and \emph{DeepMon} both have very small accuracy drop ($\leqslant$ \textbf{3\%} for top-1 and $\leqslant$ \textbf{1.5\%} for top-3).
These loss is acceptable given the observation that our baseline (no-cache) can achieve round 62.8\% top-1 accuracy and 76.1\% top-3 accuracy.
We have even observed cases where the baseline wrongly classifies the image while our \framework does it correctly.
This is because that we have designed our image matching algorithm to carefully choose which part of computations to reuse, and this reusable information is properly propagated during inference, thus minimizing the impact on the recognition output.

\begin{figure}[t]
	\centering
	\includegraphics[width=0.44\textwidth]{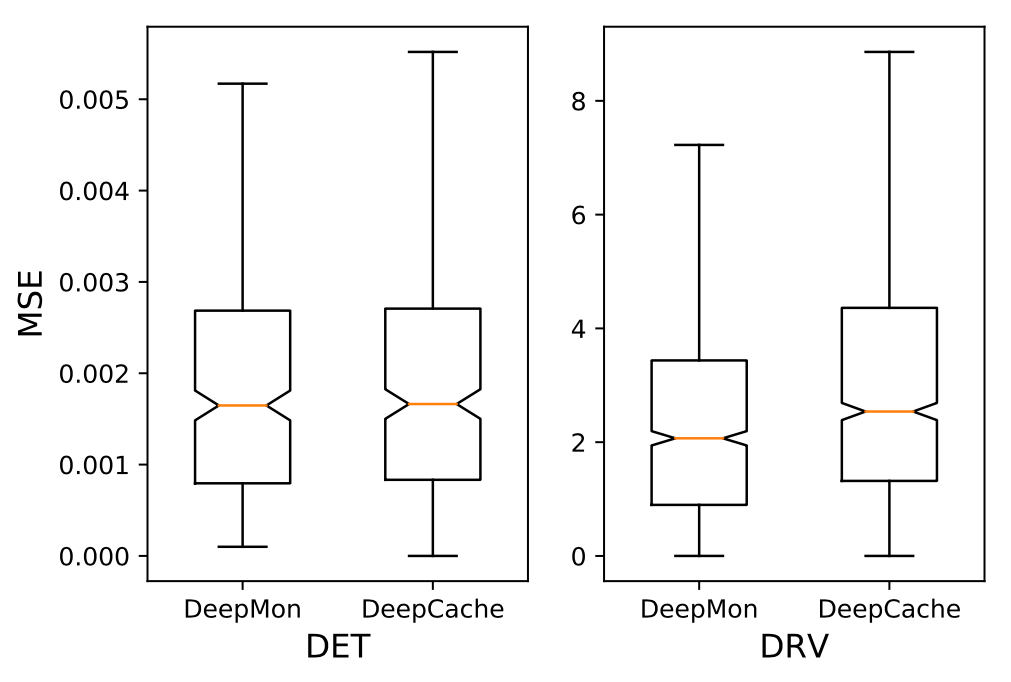}
	\caption{MSE between the output of caching approaches (\framework, \emph{DeepMon}) and ground truth (\emph{no-cache}).}
	\label{fig:eval_ed_accuracy}
\end{figure}

Figure~\ref{fig:eval_ed_accuracy} shows the MSE between ground truth (\emph{no-cache}) and other cache approaches (\framework and \emph{DeepMon}) when running DET and DRV models.
As observed, the median MSE of \framework is \textbf{0.00166} and \textbf{2.617} for DET and DRV respectively, quite similar to the results of \emph{DeepMon} with \textbf{0.00164} and \textbf{2.017}.
For the DRV case, the results can be interpreted that \framework leads to 2.6 degrees offset from the decision made by human driver.
Considering that \framework will periodically run the total image without cache reuse, as mentioned in Section~\ref{sec:overview}, this offset will not be accumulated.
To be compared, our above latency experiment shows that \framework can accelerate CNN models two times as \emph{DeepMon}, e.g., \textbf{18.2\%} vs. \textbf{8.9\%} on average across all models and benchmarks.

\subsection{Energy Saving}

\begin{figure}[t]
	\centering
	\includegraphics[width=0.44\textwidth]{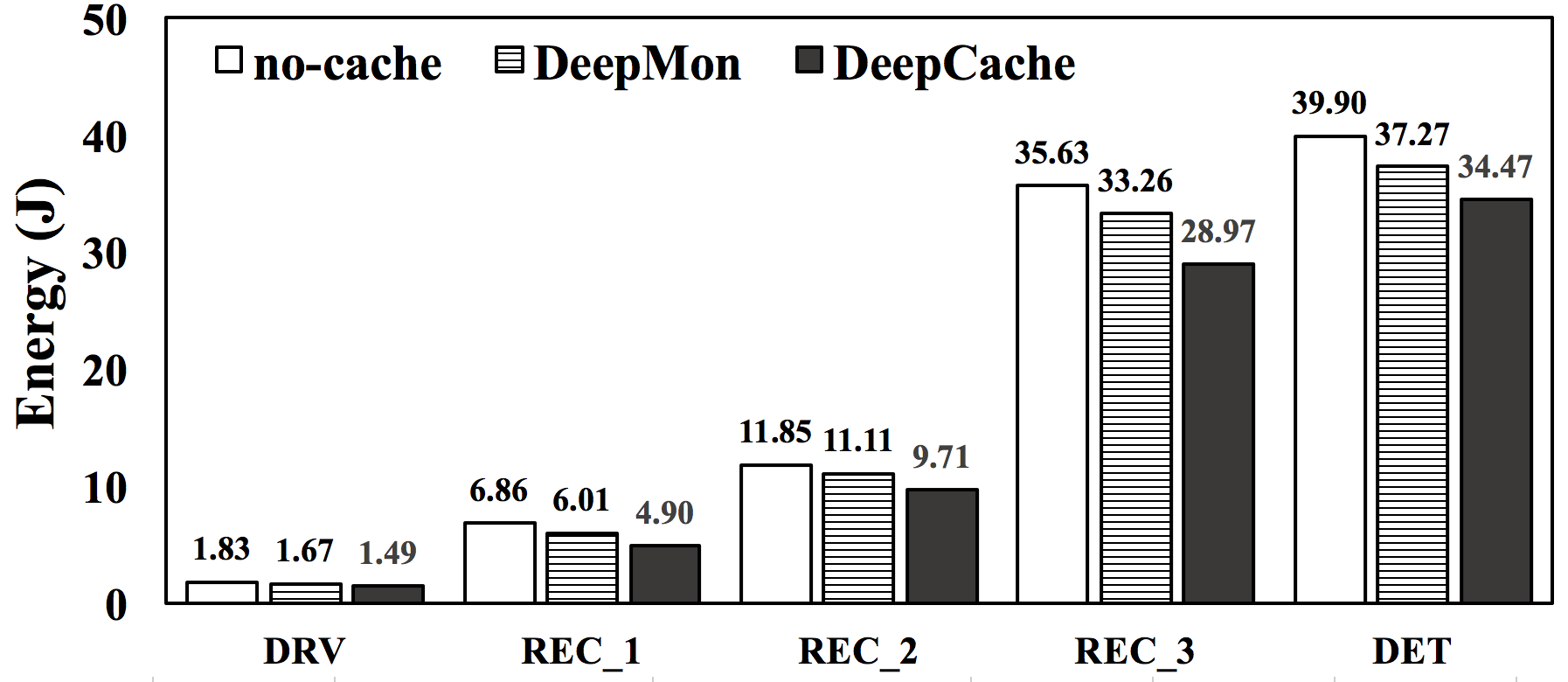}
	\caption{Energy consumption of \framework.}
	\label{fig:eval_energy}
\end{figure}

We now investigate the energy consumption of \framework across all selected benchmarks, and illustrate results in Figure~\ref{fig:eval_energy}.
It is observed that \framework can save \textbf{19.7\%} of energy consumption on average and up to \textbf{28.6\%} (REC\_1), while \emph{DeepMon} only has \textbf{8.0\%} on average.
This saving is mostly from the reduced processing time.
Considering that vision tasks are very energy-intensive, this saving up is able to substantially lengthen battery life.
For example, applying \framework on REC\_3 to classify 10 images can help spare 66.8J energy, enough to support 40 seconds of video playing on Nexus 6 phone according to our measurement.

\subsection{Choices of Parameters}\label{sec:eval_choosing}

\begin{figure}[t]
\centering
\scriptsize
\includegraphics[width=0.44\textwidth]{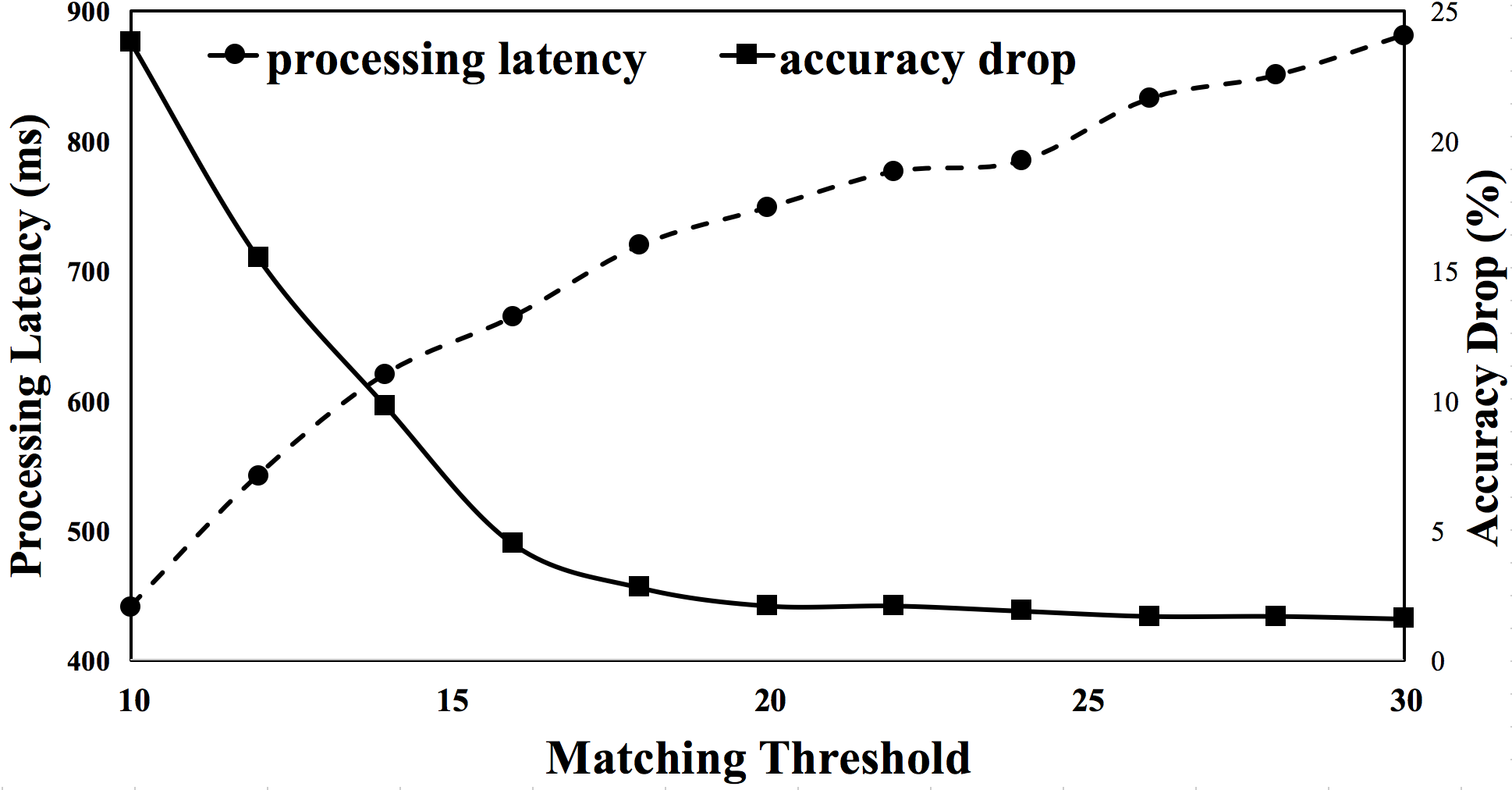}
\caption{Effect of varied matching threshold $\mathcal{T}$ on processing latency and top-1 accuracy drop of REC\_2 model.}
\label{fig:eval_parameter_T}
\end{figure}

\begin{figure}[t]
\centering
\scriptsize
\includegraphics[width=0.44\textwidth]{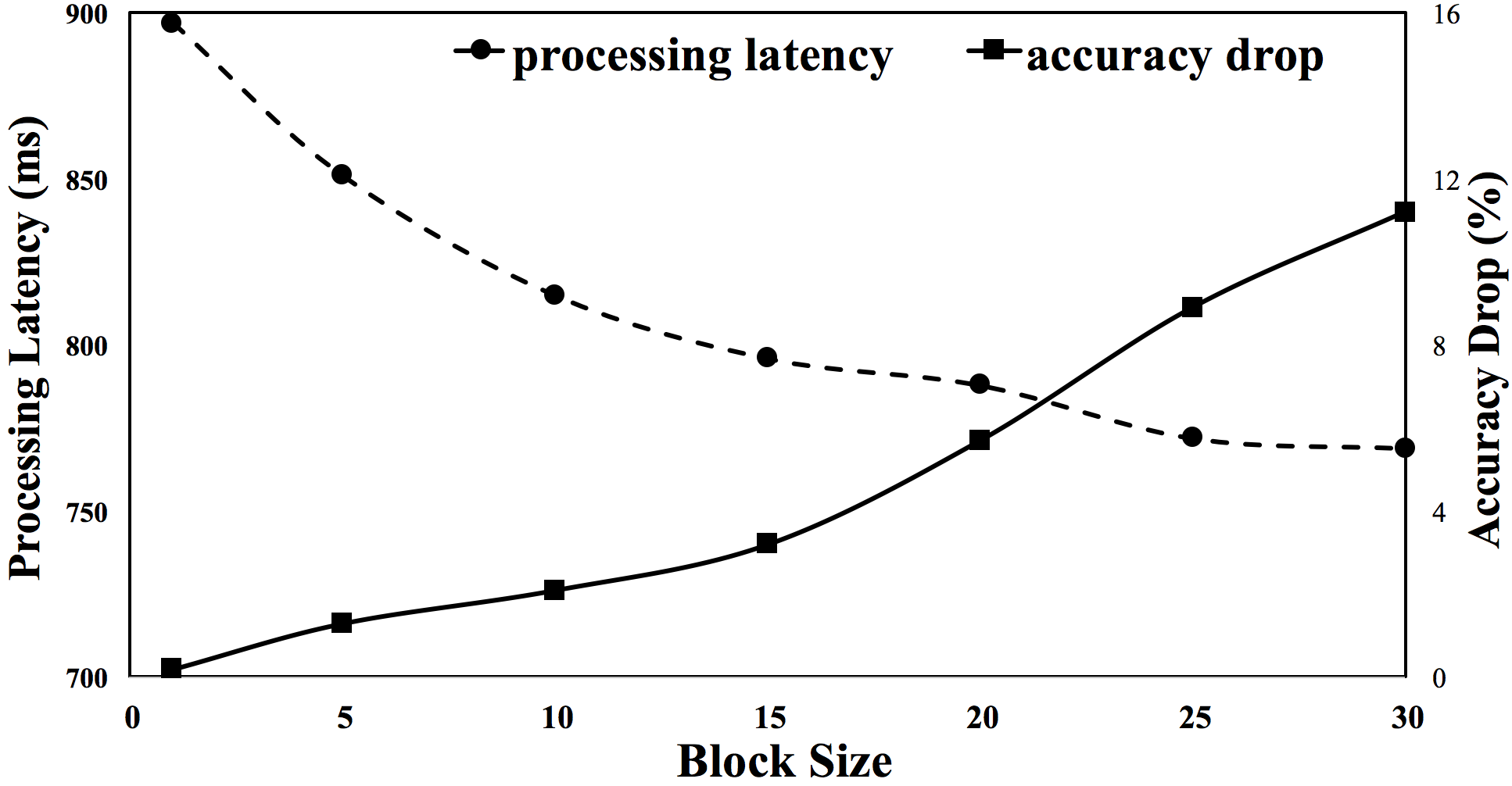}
\caption{Effect of varied block size on processing latency and top-1 accuracy drop of REC\_2 model.}
\label{fig:eval_parameter_N}
\end{figure}

In our matching algorithm mentioned in Section~\ref{sec:matching}, some variables can be used to make trade-off between latency improvement and accuracy drop.
Matching threshold $\mathcal{T}$ is the key to decide whether two image blocks are similar enough to be reused.
Figure~\ref{fig:eval_parameter_T} illustrates how $\mathcal{T}$ can affect the latency and accuracy (REC\_2 + \textbf{T1}).
As expected, higher $\mathcal{T}$ indicates fewer blocks can be matched, thus leading to less top-1 accuracy drop, but also higher processing latency.
In our default setting ($\mathcal{T} = 20$), \framework can achieve considerable latency improvement, e.g., \textbf{18.3\%} (from \textbf{917ms} to \textbf{748ms}), with acceptable accuracy loss (\textbf{2.1\%}).
This setting aligns with the fact that the acceptable values for wireless transmission quality loss are commonly considered to be about 20 to 25~\cite{psnr}.
However, the threshold can also be set by application developers to adapt to task-specific requirements.
For applications that are not very sensitive to the output accuracy, developers can aggressively use a smaller $\mathcal{T}$ to achieve higher latency improvement.

Another configurable parameter in our image matching algorithm is the block size.
As observed from Figure~\ref{fig:eval_parameter_N}, a larger block size results in more latency improvement but also higher accuracy loss.
This result is reasonable since splitting an image into large blocks indicates more coarse-grained matching.
As an extreme case, when block size equals to 1, the accuracy loss is very small (\textbf{0.2}) but the latency improvement is also very low (\textbf{2.19\%}).
This is actually the \emph{pixel-wise} approach discussed previously in Section~\ref{sec:overview}, and the result is consistent with our discussion.
Our empirical suggestion is setting block size around 10 for 227x227 images.

\subsection{Image Matching Performance}\label{sec:eval_matching}


\begin{table}[t]
\small
\centering
\begin{tabular}{L{2.5cm}ll} \hline
& \textsc{Latency (ms)} & \textsc{Match Ratio (\%)}\\\hline
DeepMon & 4.7 $\pm$ 0.7 & 46.1 \\ 
ES & 33.3 $\pm$ 13.16 & 71.5\\ 
TSS & 24.7 $\pm$ 9.41 & 70.8\\ 
DS & 19.5 $\pm$ 6.53 & 71.2\\ 
DS + optimization & 9.7 $\pm$ 2.55 & 69.5\\ \hline
\end{tabular}
\caption{\revise{A comparison of image matching algorithms between DeepMon~\cite{conf/mobisys/LocLB17} (row 1), which uses histogram-based matching, and \sys{} (the remaining rows) that uses different block matching algorithms in combination with optimization techniques mentioned in Section~\ref{sec:matching}.
\sys{} achieves much higher match ratios with minor increase in latency.}
}
\label{tab:eval_matching}
\end{table}

\revise{
Finally, we report the performance of our renderscript-based implementation of image matching algorithm individually.
Our current matching algorithm mentioned in Section~\ref{sec:matching} is based on the diamond search (DS), i.e., DS as an ``algorithm unit'' (used in Step 2). In addition to the DS, there are several other block matching algorithms that can be plugged into our image matching algorithm to replace DS, such as the Exhaustive Search (ES) and the Three Step Search (TSS). The details and differences of these algorithms are summarized in the survey effort~\cite{barjatya2004block}. In this part of evaluation, we also implement the ES-based and the TSS-based image matching to compare.
We run preceding algorithms on 10,000 images that are randomly picked from UCF101 and resized to 227x227, and log the processing time (\textit{latency}) and the proportion of matched regions (\textit{match ratio}).

As shown in Table~\ref{tab:eval_matching}, our image matching algorithm can achieve around 70\% match ratio.
The use of different block matching algorithms has minor impacts on the match ratio, but the DS-based implementation is much faster than the ES-based and TSS-based implementation, i.e., 19.5ms vs. 33.3ms \& 24.7ms.
Another important observation is that the acceleration techniques mentioned in Section~\ref{sec:matching}, i.e., k-skip and reusing, can significantly improve the processing latency from \textbf{19.5ms} to \textbf{9.7ms} on average, with only \textbf{2.4\%} loss in the match ratio.
These results indicate that our image matching algorithm works well for our CNN cache mechanism, as it occurs quite negligible overhead ($\leqslant$ \textbf{10ms}) compared to the benefit gained from cache reusing.
To be compared, the histogram-based matching algorithm used in \emph{DeepMon} matches only 46.1\% of image areas, while only runs 5ms faster.
}

In our above experiments, we treat the image matching and CNN inference as two sequential stages so that the time consumed on the image matching diminishes the benefits gained from cache-reuse.
Though the matching algorithm is accelerated, it still has non-trivial impacts on the performance of \framework especially when the model is relatively small such as DRV.
But in practice, these two stages can often be carried out asynchronously when the images can be captured at a higher rate than our CNN inference.
More specifically, \framework can run the image matching algorithm on $i$-th image and CNN inference on $(i+1)$-th image at the same time.
In our case, since we implement these two stages on different mobile processors (GPU and CPU), their processing should not interfere each other, therefore \framework can further improve the overall performance.

\subsection{Memory Overhead}

\begin{figure}[t]
	\centering
	\includegraphics[width=0.45\textwidth]{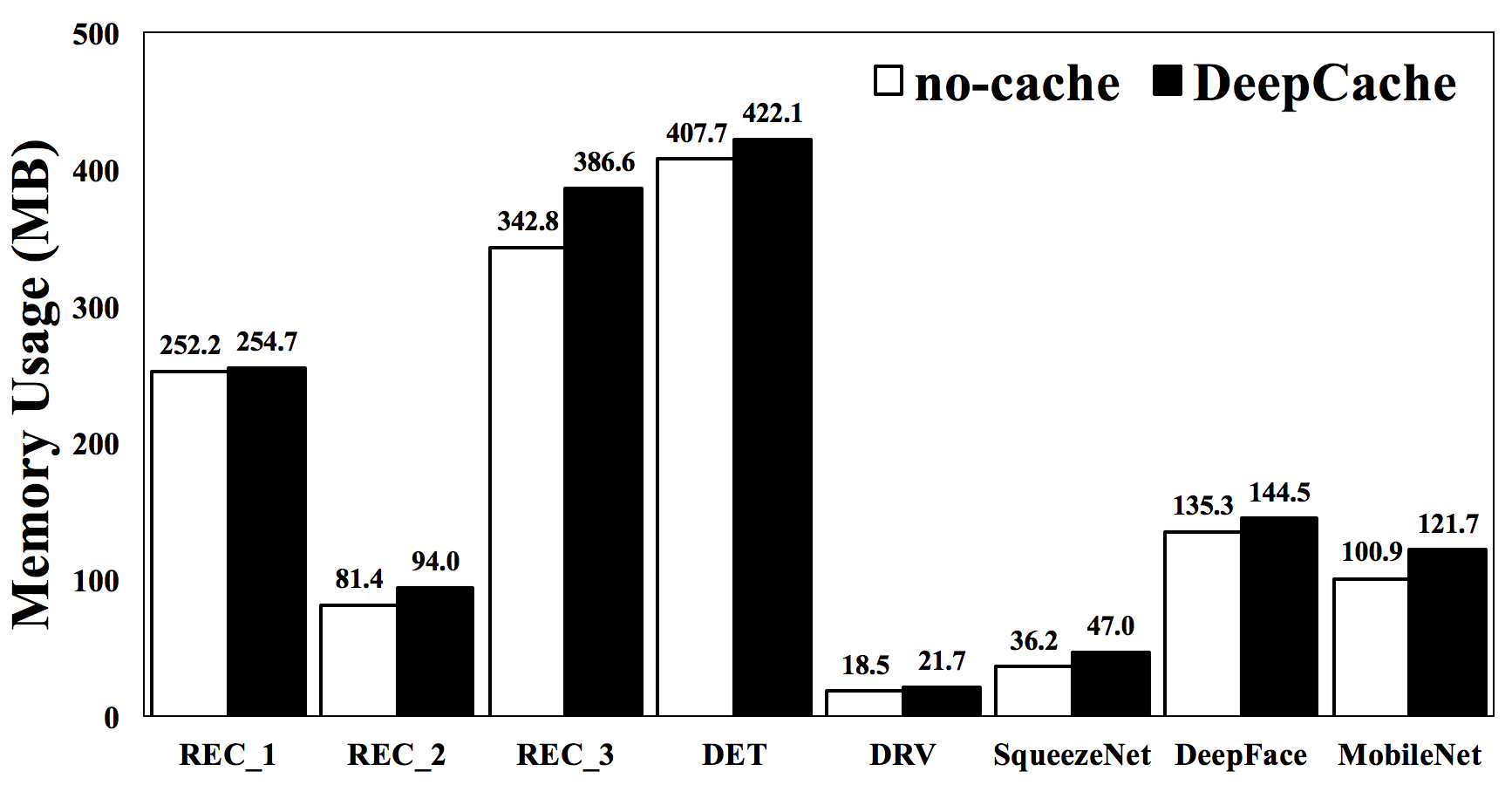}
	\caption{Memory overhead of \framework.}
	\label{fig:memory_overhead}
\end{figure}

Figure~\ref{fig:memory_overhead} shows the memory overhead of \framework.
Besides the 5 models used above, we also test on other three popular CNN models: MobileNet~\cite{MobileNet}, SqueezeNet~\cite{squeezenet}, and DeepFace~\cite{DeepFace}.
Here we assume that all model parameters are read into memory once initialized without I/O transmission during the inference.
We report the memory peak usage during the inference here.
As observed, the memory overhead occurred by \framework ranges from \textbf{2.5MB} to \textbf{43.8MB} depending on the internal structure of models.
This overhead is quite trivial since nowadays mobile devices are usually equipped with large size of memory, e.g., 3GB in Nexus 6.
Note that we only cache and reuse the computation results of convolutional layers so that no extra memory usage will be wasted on other computation-light layers.
\section{Discussion} \label{sec:discuss}
\revise{

\paragraph{Applicability to other CNN models}
This paper reports only the results of \framework on five typical CNN models.
Yet, we expect that \framework{} applies to emerging CNN models, such as the SqueezeNet~\cite{squeezenet}, the MobileNet~\cite{MobileNet}, and the DenseNet~\cite{huang2017densely}. 
Intuitively, the new models, with their innovated inter-layer organizations and intra-layer optimizations, still preserve the temporal locality that \framework hinges on.
Furthermore, our observation on the domaninating cost of early convolutional layers (Section~\ref{sec:back}) is true for these new models.

\paragraph{Implementation on accelerators}
While we prototype the inference stage of \framework{} on CPU, we expect that it can be ported to and benefit from hardware accelerators. 
Taking GPU as an example, \framework{} is capable of reducing the redundant processing by avoiding GPU kernels for computing output feature maps. 
For FPGA, we expect that our caching mechanism can be implemented as the hardware logic for further speedup. 

\paragraph{Applicability to other video types}
The idea and high-level design of \sys{} can be applied on other non-mobile videos as long as there's redundancy between adjacent frames. Currently \sys{} is optimized for the mobile vision, because (1) mobile videos contain much richer temporal locality than other video types such as edited movies, and (2) mobile devices are much more sensitive to the latency and the energy consumption in deep vision as compared to other platforms such as desktops or servers.

}


\section{Related Work}\label{sec:related}



\paragraph{Convolutional Layer Caching}
As most related efforts, DeepMon~\cite{conf/mobisys/LocLB17} and CBinfer~\cite{cavigelli2017cbinfer}  incorporate CNN caches that we deem ad-hoc.
First, they match the image blocks (or pixels) in only the same positions, therefore are unable to tolerate the scene variation as we highlighted in Section~\ref{sec:intro}.
By contrast, \sys{} retrofits proven video techniques to systematically search for nearby similar image blocks.
Second, 
they execute cache lookup over feature maps at all layers.
\revise{Such each-layer matching strategy not only incurs too much runtime overhead, but also requires extra efforts from application/model developers to manually set a ``proper'' matching threshold for each layer.
By contrast, \sys{} runs lookup only once at the input raw images, and propagates the reusable region boundaries across all the layers.
In a concurrent project, $EVA^2$~\cite{buckler2018eva} proposes hardware optimization for exploiting temporal redundancy in live computer vision. By contrast, \sys{} is designed and implemented to run on general-purpose processors that are widely available on commodity mobile devices.
Besides, $EVA^2$ requires a model to be manually separated into two parts, and the output of the prefix part will be saved and reused while the suffix part will be fully executed. In \sys{}, such manual efforts are naturally avoided by our propagation mechanism mentioned in Section~\ref{sec:cache}.
Potluck~\cite{guo2018potluck} enables the cross-application cache reuse of computations on a similar video input.
However, unlike \sys{} that identifies which parts of image regions shall be reused, the cache mechanism of Potluck is rather coarse-grained since it can reuse \textbf{only} the whole output.
}

\paragraph{Continuous Mobile Vision}
Emerging mobile vision systems span from commercial products~\cite{GoogleTranslate,amazon} to research prototypes~\cite{conf/ica3pp/OuLSE17,conf/mobisys/ZhuYZZ17,conf/cvpr/DasDWWSS17,conf/huc/HodgesWBISBSKW06,conf/mobisys/ZengCZ17,conf/mobisys/JainMC15,hwang2017raven}.
To optimize mobile vision tasks,
\cite{conf/mobisys/LiKamWaPPZB13,likamwa2013energy} made the early energy characterization and optimization towards continuous mobile vision.
Starfish~\cite{conf/mobisys/LiKamWaZ15} allows concurrent vision applications to share computation and memory objects.
RedEye~\cite{conf/isca/LiKamWaHGPZ16} reduces image sensor energy by offloading CNN layers to analog domain. 
DeepEye~\cite{conf/mobisys/MathurLBBFK17} enables rich analysis of images in near real-time via novel, small form factor wearable camera. 
Such high interest in mobile vision motivates \sys{}. 

\paragraph{Optimizing Deep Learning Execution for Mobile}
Extensive work is done on making deep learning affordable on mobile devices.
The approaches include making models much smaller to fit mobile devices~\cite{conf/huc/LaneGQ15,conf/icassp/ChenPH14,conf/icassp/VarianiLMMG14,MobileNet},
specializing hardware to deep learning algorithms~\cite{conf/asplos/ChenDSWWCT14,conf/fpga/ZhangLSGXC15,conf/isca/ChenES16,conf/isca/HanLMPPHD16},
compressing existing models~\cite{conf/mobisys/KatevasLPS17,conf/ipsn/LaneBGFJQK16,conf/cvpr/WuLWHC16,conf/nips/DentonZBLF14,conf/mobisys/HanSPAWK16,yao2017deepiot}, etc.
Complementary to these techniques, \framework speeds up mobile deep vision through systematically exploiting temporal locality in input data, across multiple inference tasks.
\sys{} can coexist with these techniques in one engine.

\section{Conclusions} 
\label{sec:conclusion}


To conclude our paper, we have proposed \framework, a principled cache design, to accelerate the execution of CNN models via leveraging video temporal locality for continuous vision tasks.
At the beginning of model input, \framework discovers temporal locality by exploiting the video's internal structure, for which it borrows proven heuristics from video compression;
into the model, \framework propagates reusable result regions by exploiting the model's internal structure.
We have implemented a prototype of \framework to run unmodified CNN models on commodity Android device, and comprehensively evaluate its effectiveness via a set of experiments on typical CNN models.
\section*{Acknowledgment}
This work was supported by the National Key R\&D Program under the grant number 2018YFB1004801, the National Natural Science Foundation of China under grant numbers 61725201, 61528201, 61529201, and a Google Faculty Award. 

\bibliographystyle{ACM-Reference-Format}
\balance
\bibliography{secs/ref}

\nocite{liu2007towards}

\end{document}